\documentclass{article} 
\usepackage[margin=1in]{geometry}
\usepackage{times}
\usepackage{natbib}

\usepackage{amsmath,amsfonts,bm}

\def\eqref#1{equation~\ref{#1}}

\def\1{\bm{1}}

\DeclareMathAlphabet{\mathsfit}{\encodingdefault}{\sfdefault}{m}{sl}
\SetMathAlphabet{\mathsfit}{bold}{\encodingdefault}{\sfdefault}{bx}{n}

\newcommand{\KL}{D_{\mathrm{KL}}}

\usepackage{hyperref}
\usepackage{url}
\usepackage{graphicx}
\usepackage{booktabs}
\usepackage{placeins}
\usepackage{wrapfig}
\usepackage{amsmath}
\usepackage{amssymb}
\usepackage{amsthm}
\usepackage{mathtools}
\usepackage{bm}

\newtheorem{theorem}{Theorem}
\newtheorem{lemma}{Lemma}
\newtheorem{proposition}{Proposition}

\newcommand{\method}{OSOL}
\newcommand{\methodmath}{\mathrm{OSOL}}
\newcommand{\bestscore}[1]{\textbf{#1}}
\newcommand{\secondscore}[1]{\underline{#1}}

\newcommand{\score}{\bm{\varphi}}
\newcommand{\Phimat}{\bm{\Phi}}
\newcommand{\Kmat}{\bm{K}}
\newcommand{\Fmat}{\bm{F}}
\newcommand{\Fbar}{\overline{\bm{F}}}
\newcommand{\bdelta}{\bm{\delta}}

\newcommand{\bxi}{\bm{\xi}}
\newcommand{\bh}{\bm{h}}

\title{One Step, One Lead: Mitigating Higher-Order Interference in Multi-Domain Reinforcement Learning via Cross-Step Control}

\author{Zihan Lin$^{1,2,3}$\thanks{Both authors contributed equally to this research.}\hspace{0.4em}\thanks{Work was done during an internship at Meituan.}\hspace{0.4em}, Xiaohan Wang$^{2}$\footnotemark[1] \thanks{Corresponding authors: \texttt{wangxiaohan17@meituan.com},\texttt{yinguojun02@meituan.com}, \texttt{ran.he@ia.ac.cn}}\hspace{0.4em}, Jie Cao$^{1}$, Jiajun Chai$^{2}$, \\
{Guojun Yin$^{2}$\footnotemark[3]\hspace{0.4em}, Wei Lin$^{2}$, Ran He$^{1}$\footnotemark[3]} \\
\\
$^1$MAIS\&NLPR, Institute of Automation, Chinese Academy of Sciences \\
$^2$Meituan, Beijing \\
$^3$School of Advanced Interdisciplinary Sciences, University of Chinese Academy of Sciences
}

\date{}
\begin{document}

\maketitle

\begin{abstract}

Reinforcement learning (RL) across multiple domains can broaden the reasoning capabilities of large language models (LLMs), yet joint training often degrades individual-domain performance and can destabilize optimization. Existing work typically diagnoses such interference from a single-step view using first-order gradient alignment or curvature-based proxies. We show that this view can miss a critical form of sequential interference: same-point domain gradients may remain nearly orthogonal even when consecutive realized updates partially reverse one another in output space. We further show that consecutive token log-probability footprints recover this interaction directly from adjacent checkpoints as a local second-order interaction in output space, without explicitly reconstructing same-step curvature. Building on this insight, we propose \method, which designates a focus domain at each iteration, uses the preceding checkpoint footprint to rank token-level rebound risk, and applies a drift-ranked, adaptively scaled correction within the standard GRPO update. Our analysis shows that this correction suppresses the targeted cross-step output backtracking component. Controlled studies further show that cross-step backtracking is more strongly associated with subsequent task damage than same-point gradient diagnostics, while the preceding footprint ranks future rebound risk more accurately than Hessian-based proxies. On Qwen3-30B-A3B, \method\ reaches a domain-macro average of $0.4822$, improving by $5.7\%$ over the strongest compared baseline, without explicit higher-order differentiation.

\end{abstract}





\section{Introduction}
\label{intro}

\begin{figure}[t]
    \centering
    \includegraphics[width=0.94\linewidth]{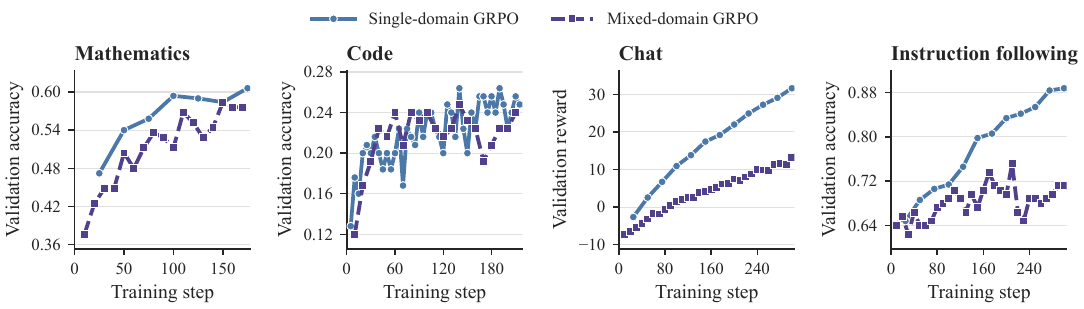}
    \caption{
    \textbf{Mixed-domain GRPO leaves substantial domain-specific headroom relative to isolated training.}
    Validation trajectories of Qwen3-4B trained with GRPO either separately on each domain or jointly on the four-domain mixture.
    Clear gaps emerge in Mathematics, Chat, and Instruction Following, with a smaller separation in Code, highlighting the domain dependence of joint-training degradation.
    }
    \label{fig:single-vs-mixed}
    \vspace{-6pt}
\end{figure}

Reinforcement learning (RL) has been widely used to improve the reasoning capabilities of large language models (LLMs) in individual domains \cite{shao2024deepseekmath,lin2026resrlboostingllmreasoning,yu2026dapo}. However, joint training across heterogeneous domains often yields uneven domain-wise gains and can even destabilize training \cite{cai2026advancing,yang2026domains,li2025can}. The resulting degradation can be strongly domain dependent, as illustrated by the gap between mixed-domain and single-domain GRPO in Figure~\ref{fig:single-vs-mixed}. Such interference limits the ability of RL to improve a broad set of capabilities within a single foundation model. We investigate how mixed-domain updates interact along the training trajectory and how this interference can be controlled without introducing auxiliary expert models or reward signals beyond those already used by the underlying RL training \cite{lin2026rest,liu2026cdrrm}.

\begin{figure}[t]
  \centering
  \includegraphics[width=0.94\linewidth]{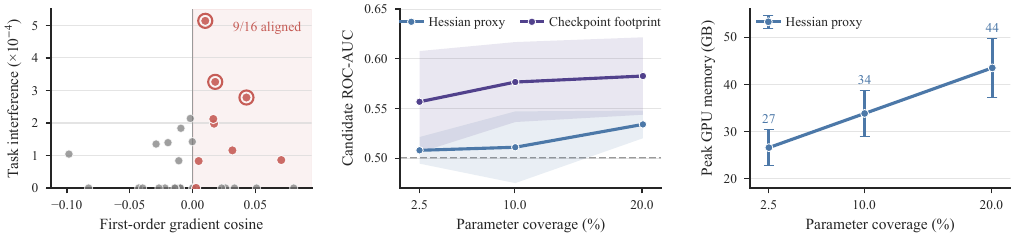}
\caption{\textbf{Same-point alignment misses sequential damage, while broader curvature estimation remains costly.} (a) Positive same-point gradient cosine does not preclude substantial task damage after the subsequent domain update; circles highlight high-damage aligned cases. (b, c) Expanding Hessian-proxy parameter coverage modestly improves token-level rebound identification but increases peak GPU memory.}
  \label{fig:hessian-coverage-cost}
  \vspace{-4pt}
\end{figure}

Explaining this domain-dependent degradation requires identifying which interactions among heterogeneous updates are associated with subsequent capability degradation. Existing approaches mainly analyze and control such interference within a single training step, either by comparing first-order gradients or by estimating higher-order curvature \cite{yu2020gradient,liu2021conflict,yang2026local}. However, our controlled study shows that same-point first-order gradient alignment does not reliably reflect the task damage caused by the subsequent update, as shown in Figure~\ref{fig:hessian-coverage-cost}(a). Nine of 16 replicates have positive gradient cosine (i.e., are aligned), yet aligned cases can still incur substantial task damage. This result indicates that same-point alignment alone is insufficient for diagnosing sequential interference. This limitation has motivated higher-order approaches, which use curvature to estimate how one domain update changes the effect of another \cite{liang2026boosting,yang2026local}. However, expanding the parameter coverage of a Hessian proxy yields only modest gains in token-level rebound identification while substantially increasing peak GPU memory, revealing an unfavorable trade-off between estimation quality and computational cost, as shown in Figure~\ref{fig:hessian-coverage-cost}(b,c). The limited improvement under broader coverage suggests that increasing parameter coverage within a local curvature approximation recovers only limited additional information relative to the realized checkpoint trajectory. Together, these findings motivate us to move beyond gradient and curvature estimates constructed within a single step and instead examine the actual token log-probability footprints left by consecutive updates along the training trajectory.


Building on this cross-step view, we show that adjacent token footprints reveal how consecutive updates interact in task-conditioned output space and provide a principled signal for reducing the resulting cross-step output backtracking. This motivates OSOL, which designates a focus domain at each iteration and converts the preceding token footprint into fine-grained history corrections within the standard GRPO update. These corrections are derived directly from the token-level policy shift left by the preceding update, requiring neither auxiliary supervision nor explicit higher-order derivatives. On Qwen3-30B-A3B, \method\ reaches a domain-macro average of $0.4822$, improving by $5.7\%$ over MGS, the strongest baseline, while also outperforming CGPO. The main contributions of this paper are summarized as follows:

1) We introduce a cross-step view of higher-order interference in multi-domain policy optimization, shifting the analysis from same-point gradient relations and the curvature sensitivity of isolated parameter displacements to the mixed interaction between consecutive updates in task-conditioned output space. We prove that the inner product of the token log-probability footprints induced by the preceding update and the current base update locally recovers this interaction under the empirical output Fisher, making it observable from adjacent checkpoints without explicitly constructing the Fisher matrix. We further show that opposing token movements give this interaction a direct trajectory interpretation as cross-step output backtracking.

2) Building on this analysis, we propose \method\, an online cross-step controller integrated into the standard policy gradient update. At each iteration, it designates a focus domain, forms a normalized task-weighted base coefficient, and computes token-level drift from the log-probability difference of the response tokens across adjacent checkpoints; negative drift defines the intervention set, its rank determines relative penalty strength, and adaptive scaling keeps the residual at a controlled fraction of the current base-coefficient dispersion. The resulting residual is added directly to the objective, and we prove that it contracts the targeted negative-to-positive backtracking component under a local response condition.

3) Empirical evaluations validate both the diagnostic and optimization benefits of the cross-step view. In controlled studies, cross-step backtracking mass correlates positively with subsequent task damage ($\rho_s=0.40$, 95\% bootstrap CI $[0.05,0.69]$), whereas same-point global and module diagnostics yield correlations of $-0.19$ and $-0.22$, respectively. Checkpoint-footprint ranking also improves rebound ROC-AUC from $0.498$ for the Hessian proxy to $0.567$. In full multi-domain training on Qwen3-30B-A3B, \method\ improves the domain-macro average by $5.7\%$ over MGS, the strongest compared baseline, while also outperforming curvature-guided CGPO. Ablations further show that equalizing the focus and non-focus weights or replacing drift-ranked token weights with random weights reduces the domain-macro average by $51.4\%$ and $50.8\%$, respectively.

\section{Related Work}
\label{relate}
Recent studies of RL post-training across heterogeneous reasoning domains report asymmetric transfer, domain degradation, task-dependent gradient imbalance, and sensitivity to data composition and training order \citep{cheng2026revisiting,li2025can,wu2026imbalanced,yang2026domains}. Existing optimization methods mainly balance task contributions or coordinate gradients evaluated at the same checkpoint. Adaptive weighting methods balance gradient magnitudes or task convergence rates \citep{chen2018gradnorm,gong2024coba}, while multi-objective methods seek Pareto-efficient, bargaining-based, or balanced-progress solutions \citep{sener2018multi,navon2022multi,liu2023famo}. PCGrad and CAGrad instead modify conflicting task gradients before aggregation \citep{yu2020gradient,liu2021conflict}. Recent LLM-specific RL methods instantiate these ideas through module-level gradient surgery, adaptive task weighting and sampling, and transfer-aware curricula \citep{cai2026advancing,ramesh2026multi,yang2026transferability}. These approaches primarily regulate task contributions or same-step gradient interactions, but do not directly model interference between consecutive mixed-domain updates.

Characterizing such cross-step interaction connects three relevant perspectives: second-order sensitivity, output-space policy geometry, and checkpoint-based observability. Influence functions estimate sensitivity to infinitesimal training perturbations through inverse-Hessian information \citep{koh2017understanding}. Recent local-perturbation theory models cross-domain damage through the curvature of an earlier-domain objective under a later displacement \citep{yang2026local}, while CGPO uses cross-domain curvature information to shape multi-domain policy updates \citep{liang2026boosting}. Natural-gradient and trust-region methods provide a complementary policy-geometry view, where the Fisher metric and policy KL quantify local changes in the action distribution \citep{amari1998natural,kakade2001natural,schulman2015trust,martens2020new}. Work on the empirical Fisher cautions against treating it as a generic substitute for an arbitrary objective Hessian \citep{kunstner2019limitations}. Checkpoint-based influence estimation further shows how stored checkpoints can expose training effects along the optimization trajectory \citep{pruthi2020estimating}. Existing curvature-based analyses typically characterize isolated update sensitivity and can be costly to apply online at LLM scale. In contrast, we focus on the directly observable interaction between consecutive updates in task-conditioned output space.

\section{Cross-Step Dynamics}
\label{sec:theory}

We use a \emph{token log-probability footprint} to denote the vector of log-probability changes that a parameter update induces on a fixed set of realized response-token coordinates. Each coordinate $(x_i,a_i)$ pairs a realized token $a_i$ with its prefix context $x_i$. Accordingly, task-conditioned output space refers here to the vector of token log-probabilities evaluated on such coordinates from the domain being analyzed. We now formalize the mixed output interaction between the preceding realized update and the current base update in this output space. Consider
\begin{equation}
\label{eq:cross_updates}
\bm\theta_{k-1}\xrightarrow{\;\bdelta_{k-1}\;}\bm\theta_k
\xrightarrow{\;\bdelta_k^0\;}\bm\theta_{k+1}^0,
\end{equation}
where $\bdelta_{k-1}:=\bm\theta_k-\bm\theta_{k-1}$ is the preceding realized update and $\bdelta_k^0:=\bm\theta_{k+1}^0-\bm\theta_k$ is the current base update before the \method\ correction. Focusing on the mixed output interaction between adjacent updates makes the relevant geometry checkpoint-observable, token-resolved, and amenable to online control.

\subsection{Output Geometry}
\label{sec:cross_step_geometry}

Let $\mathcal E_k=\{(x_i,a_i)\}_{i=1}^N$ denote eligible response-token coordinates from the domain selected for cross-step control at iteration $k$, which we call the \emph{focus domain}; Section~\ref{sec:algorithm} gives its role in the training update and Section~\ref{sec:method_history} specifies the implementation-level eligibility rule. Let
$\bm\ell_k(\bm\theta):=[\log\pi_{\bm\theta}(a_i\mid x_i)]_{i=1}^N$ collect the corresponding token log-probabilities. Define the output KL on these focus-domain token contexts
\begin{equation}
\label{eq:output_divergence}
\mathcal D_k(\bdelta)
:=\frac1N\sum_{i=1}^N
\KL\!\left(\pi_{\bm\theta_k}(\cdot\mid x_i)\,\Vert\,
\pi_{\bm\theta_k+\bdelta}(\cdot\mid x_i)\right).
\end{equation}
For $\score_{i,a}:=\nabla\log\pi_{\bm\theta_k}(a\mid x_i)$ and $\score_i:=\score_{i,a_i}$, define
\begin{equation}
\Fbar_k:=\frac1N\sum_i\mathbb E_{a\sim\pi_{\bm\theta_k}(\cdot\mid x_i)}[\score_{i,a}\score_{i,a}^\top],\qquad
\Fmat_k:=\frac1N\sum_i\score_i\score_i^\top.
\end{equation}
The first, which we call the output Fisher, averages score outer products over the full next-token distribution at each context. The second is the empirical output Fisher obtained from the response tokens actually realized in the rollout \citep{amari1998natural,martens2020new,kunstner2019limitations}. This distinction lets the KL geometry be defined over the policy distribution while keeping the cross-step interaction observable on realized tokens.

\begin{lemma}[Task-conditioned output Hessian]
\label{lem:output_hessian}
Under the local regularity conditions in Appendix~\ref{app:standing},
\begin{equation}
\label{eq:output_hessian}
\mathcal D_k(\bm0)=0,\quad
\nabla\mathcal D_k(\bm0)=\bm0,\quad
\nabla^2\mathcal D_k(\bm0)=\Fbar_k,\quad
\mathcal D_k(\bdelta)=\tfrac12\bdelta^\top\Fbar_k\bdelta+O(\|\bdelta\|^3).
\end{equation}
\end{lemma}
Thus the relevant geometry is the exact local Hessian of output KL---the metric underlying natural-gradient and trust-region policy optimization \citep{kakade2001natural,schulman2015trust}---rather than an identification of the complete objective Hessian with the Fisher. Proofs are in Appendices~\ref{app:kl_hessian} and~\ref{app:kl_remainder}.

On the same realized coordinates, define the preceding footprint and the current base footprint
\begin{equation}
\label{eq:cross_footprints}
\bm u:=\bm\ell_k(\bm\theta_k)-\bm\ell_k(\bm\theta_{k-1}),\qquad
\bm d^0:=\bm\ell_k(\bm\theta_{k+1}^0)-\bm\ell_k(\bm\theta_k).
\end{equation}

\begin{theorem}[Checkpoint token--Fisher duality]
\label{thm:cross_step}
If $\|\bdelta_{k-1}\|,\|\bdelta_k^0\|=O(\eta)$, then
\begin{equation}
\label{eq:cross_duality}
\frac1N\bm u^\top\bm d^0
=\bdelta_{k-1}^\top\Fmat_k\bdelta_k^0+O(\eta^3).
\end{equation}
\end{theorem}
The inner product of the preceding and current base footprints therefore exposes the mixed output interaction between the preceding realized update and the current base update under the empirical output Fisher. Both footprints are defined on the same realized token coordinates, while the online controller needs only the already available preceding footprint $\bm u$. The coordinatewise expansion and explicit finite-step error are given in Appendices~\ref{app:first_footprint}--\ref{app:token_duality}.

\paragraph{Token Backtracking.}
The interaction has the exact path interpretation
\begin{equation}
\label{eq:gamma_path}
\Gamma(\bm u,\bm d)
:=-\frac1N\bm u^\top\bm d
=\frac{\|\bm u\|^2+\|\bm d\|^2-\|\bm u+\bm d\|^2}{2N}.
\end{equation}
Hence $\Gamma>0$ measures stepwise output motion that is canceled in the net displacement. Together with Theorem~\ref{thm:cross_step},
\begin{equation}
\Gamma(\bm u,\bm d^0)=-\bdelta_{k-1}^\top\Fmat_k\bdelta_k^0+O(\eta^3),
\end{equation}
which gives the mixed output interaction a direct trajectory interpretation. Appendix~\ref{app:path_cancellation} decomposes the global quantity as $\Gamma=\mathrm{Rev}-\mathrm{Align}$, where $\mathrm{Rev}=\mathcal B^-+\mathcal B^+$ separates token reversals from aligned motion. The nonpositive history correction used by \method\ has a monotone reduction guarantee for the negative-to-positive component; see Appendix~\ref{app:penalty_asymmetry}. We therefore define the controlled backtracking mass as
\begin{equation}
\label{eq:reversal_mass}
\mathcal B^-(\bm u,\bm d)
:=\frac1N\sum_{i:u_i<0}(-u_i)[d_i]_+,
\end{equation}
which captures negative-to-positive rebound: a token log-probability decreased by the preceding update and immediately increased by the next one.

\subsection{Backtracking Control}
\label{sec:backtracking_control}

\method\ applies a nonpositive residual on
$\mathcal C_k:=\{i\in\mathcal E_k:u_i<0\}$. Let $\xi_i\leq0$ denote its effective post-clipping change to the policy coefficient; monotonicity of scalar clipping preserves this sign (Appendix~\ref{app:advantage_clipping}).

\begin{lemma}[History-conditioned PPO penalty]
\label{lem:history_penalty}
For $z_i:=\log\!\left(\pi_{\bm\theta}(a_i\mid x_i)/\pi_{\bm\theta_k}(a_i\mid x_i)\right)$, the token-level loss difference at the PPO anchor satisfies
\begin{equation}
\label{eq:main_token_penalty}
\mathcal J_i^{\methodmath}(z_i)-\mathcal J_i^0(z_i)=-e^{z_i}\xi_i,
\qquad
\left.\partial_{z_i}(\mathcal J_i^{\methodmath}-\mathcal J_i^0)\right|_{z_i=0}=-\xi_i\geq0.
\end{equation}
\end{lemma}
Thus gradient descent adds a history-conditioned penalty against increasing the selected likelihood ratios. Appendix~\ref{app:ppo_penalty} derives the result from the clipped PPO surrogate \citep{schulman2017proximal}.

Let $\Phimat_{\mathcal E}:=[\score_1,\ldots,\score_N]$ and
$\Kmat_{\mathcal E}:=\Phimat_{\mathcal E}^\top\Phimat_{\mathcal E}$, where $\Kmat_{\mathcal E}$ is the token-gradient similarity matrix and captures cross-token coupling induced by shared model parameters. In the local policy-gradient model, the correction induces
\begin{equation}
\label{eq:main_residual_footprint}
\bh:=\Phimat_{\mathcal E}^\top\bdelta_k^r
=\frac{\eta}{|\mathcal R|}\Kmat_{\mathcal E}\bxi,
\end{equation}
where $\bh$ is the correction footprint.

Because all tokens share model parameters, a negative coefficient correction does not by itself guarantee a negative log-probability change for every selected token: cross-token interactions in $\Kmat_{\mathcal E}$ can alter the local response. The following sufficient condition requires each selected token's own local response to dominate the aggregate cross-token effect.
\begin{lemma}[Residual-to-output sign transfer]
\label{lem:sign_transfer}
Under the local dominance condition in Appendix~\ref{app:self_response}, $h_i\leq0$ for the selected coordinates.
\end{lemma}
Under this condition, the nonpositive coefficient correction transfers to a nonpositive output-space correction on the selected tokens. Appendix~\ref{app:self_response} gives the exact condition and its extension to the locally preconditioned optimizer response.

\begin{proposition}[Backtracking contraction]
\label{prop:backtracking_reduction}
If $h_i\leq0$ on $\mathcal C_k$, then
\begin{equation}
\label{eq:control_gain}
\Delta_k^{\mathrm{ctrl}}
:=\mathcal B^-(\bm u,\bm d^0)-\mathcal B^-(\bm u,\bm d^0+\bh)\geq0.
\end{equation}
For the corrected footprint $\bm d^{\methodmath}$ with $\|\bdelta_k^r\|=O(\eta)$,
\begin{equation}
\label{eq:finite_control_gain_main}
\mathcal B^-(\bm u,\bm d^{\methodmath})
\leq\mathcal B^-(\bm u,\bm d^0)-\Delta_k^{\mathrm{ctrl}}+O(\eta^3).
\end{equation}
\end{proposition}
The proof and explicit finite-step remainder are in Appendices~\ref{app:backtracking_proof} and~\ref{app:finite_control}. The theory therefore closes an observable-to-controllable loop: output KL identifies the local geometry, the preceding and current base footprints expose the mixed output interaction, token signs localize backtracking, and the history-conditioned penalty contracts the negative-to-positive component targeted by \method.

\section{Method Design}
\label{sec:algorithm}

\paragraph{One Step, One Lead.}
Each mixed-domain update designates exactly one domain $f_k$ as the \emph{focus domain}; the remaining domains are \emph{non-focus domains}. The focus domain receives a larger task weight in the base GRPO coefficient, and only its eligible response tokens receive the history-based correction. The non-focus domains remain in the same update with smaller task weights, so the method prioritizes one domain without discarding the others. A training schedule selects the focus domain at each iteration.

At iteration $k$, \method\ samples the standard mixed-domain rollout batch from $\pi_{\bm\theta_k}$ and rescores the same realized eligible focus-domain tokens with the frozen preceding checkpoint $\bm\theta_{k-1}$. The resulting drift defines the intervention set ($u_i<0$), its rank determines relative strength, and $\tau$ sets the batchwise scale. The residual is then added to the existing GRPO coefficient and optimized with the unchanged clipped policy objective. Thus \method\ modifies the update signal rather than introducing a separate auxiliary loss. All history-dependent quantities are stop-gradient. The focus assignment determines where the checkpoint controller is evaluated; it does not alter the definition of the underlying group-relative reward signal.

\subsection{Update History}
\label{sec:method_history}

Let $A_i^{\mathrm g}$ be the GRPO group-normalized trajectory advantage broadcast to token $i$, and let $\mathrm{src}(i)$ denote its domain. With focus and non-focus weights $w_{\mathrm f}>w_{\mathrm{nf}}$, the task-weighted base coefficient is
\begin{equation}
\label{eq:m_taskw}
A_i^{\mathrm{base}}=c_iA_i^{\mathrm g},\qquad
c_i=\frac{w_{\mathrm{src}(i)}}{Z_k},\qquad
Z_k=\frac{w_{\mathrm f}N_k^{\mathrm f}+w_{\mathrm{nf}}N_k^{\mathrm{nf}}}
{N_k^{\mathrm f}+N_k^{\mathrm{nf}}},
\end{equation}
where $N_k^{\mathrm f}$ and $N_k^{\mathrm{nf}}$ are focus- and non-focus-token counts. The normalization keeps the coefficient scale comparable as the designated focus changes. The history residual is added after this weighting and is not multiplied by $c_i$ again.

Let $\mathcal E_k$ contain current focus-domain response tokens with nonzero trajectory-level GRPO advantage. On each realized coordinate $(x_i,a_i)$, compute
\begin{equation}
\label{eq:m_history}
u_i=\log\pi_{\bm\theta_k}(a_i\mid x_i)-
\log\pi_{\bm\theta_{k-1}}(a_i\mid x_i),\qquad
\mathcal C_k=\{i\in\mathcal E_k:u_i<0\}.
\end{equation}
Thus $\bm u$ is the preceding footprint on the current rollout coordinates, not a prediction of the next update. Both checkpoints score the same $(x_i,a_i)$ pairs, so the comparison requires no token matching or auxiliary model. A token belongs to $\mathcal C_k$ when the previous mixed-domain step reduced its current realized log-probability. This defines the negative-drift candidate set on which a nonpositive correction monotonically suppresses negative-to-positive rebound (Appendix~\ref{app:penalty_asymmetry}).

\subsection{Residual Construction}
\label{sec:method_residual}

For $i\in\mathcal C_k$, let $\mathrm{rank}_0(u_i)$ be the ascending zero-based rank and set
\begin{equation}
\label{eq:m_quant}
p_i=\frac{\mathrm{rank}_0(u_i)+\tfrac12}{|\mathcal C_k|},\qquad
\bar r_i=r_{\min}+
\frac{\mathrm{clip}(p_i,p_{\mathrm{lo}},p_{\mathrm{hi}})-p_{\mathrm{lo}}}
{p_{\mathrm{hi}}-p_{\mathrm{lo}}}(r_{\max}-r_{\min}),
\end{equation}
with $\bar r_i=0$ off $\mathcal C_k$. We use
$(p_{\mathrm{lo}},p_{\mathrm{hi}},r_{\min},r_{\max})=(0.1,1,-1,0)$, assigning stronger penalties to more negative drift. This ordering is locally optimal for allocating a fixed collection of penalty levels to active rebound coordinates (Appendix~\ref{app:rank_allocation}). Using ranks rather than raw drift magnitudes also reduces sensitivity to batch-to-batch scale changes and isolated extreme log-probability shifts.

Define the supported profile and its adaptive scale by
\begin{align}
q_i&:=\bar r_i\mathbf 1\{i\in\mathcal C_k\},
\label{eq:m_effective_raw}\\
\kappa_k&:=\tau
\frac{\mathrm{std}_{\mathcal E_k}(A^{\mathrm{base}})}
{\mathrm{std}_{\mathcal E_k}(q)},
\qquad
r_i^{\methodmath}:=\kappa_kq_i.
\label{eq:m_scale}
\end{align}
Whenever the statistics are nondegenerate,
\begin{equation}
\label{eq:m_scale_identity}
\mathrm{std}_{\mathcal E_k}(r^{\methodmath})
=\tau\,\mathrm{std}_{\mathcal E_k}(A^{\mathrm{base}}).
\end{equation}
The standard setting uses $\tau=0.03$. Hence $\tau$ controls the residual relative to the current base-coefficient scale rather than to the raw scale of the preceding drift. Numerical fallback and the code-level gate and mask are specified in Appendix~\ref{app:code_construction}--\ref{app:quantile_scale}.

The final policy coefficient is
\begin{equation}
\label{eq:m_adv}
\widetilde A_i=\mathrm{clip}\!
\left(A_i^{\mathrm{base}}+r_i^{\methodmath},-3,3\right).
\end{equation}
With $\omega_i(\bm\theta):=\pi_{\bm\theta}(a_i\mid x_i)/\pi_{\bm\theta_{\mathrm{old}}}(a_i\mid x_i)$, it is optimized by the unchanged clipped surrogate \citep{schulman2017proximal}:
\begin{equation}
\label{eq:m_obj_main}
\mathcal L_{\methodmath}(\bm\theta)
=-\mathbb E\!\left[
\frac1{|\mathcal R|}\sum_{i\in\mathcal R}
\min\!\left(
\omega_i(\bm\theta)\widetilde A_i,
\mathrm{clip}(\omega_i(\bm\theta),1-\epsilon_c,1+\epsilon_c)\widetilde A_i
\right)
\right].
\end{equation}
Thus, \method\ converts the preceding token drift into a tokenwise history-conditioned correction within the standard GRPO objective.

\section{Experiments}
\label{exp}

\subsection{Setup}
\paragraph{Models and baselines.}
We evaluate \method\ on Qwen3-30B-A3B and Qwen3-8B-Base. On both backbones, we compare against the pre-RL base model, standard GRPO~\citep{guo2025deepseek}, GRPO$+$KL$_{0.001}$ as a conservative-update baseline, MGS~\citep{cai2026advancing} as a first-order module-level gradient-surgery baseline, and CGPO~\citep{liang2026boosting} as a curvature-guided multi-domain optimization baseline. All methods follow the same evaluation protocol. Complete main-training and generation settings are summarized in Appendix~\ref{app:main_training_config}.

\paragraph{Training data.}
We construct a training dataset mixture covering four domains: mathematics, code generation, instruction following (IF), and open-ended chat (Chat). The mathematics and code examples are quality-filtered subsets of the NVIDIA Nemotron Post-Training Dataset v2~\citep{basant2025nvidia}. For Chat, we use the WildChat-IF subset provided by MGS, which is derived from the Tulu 3 WildChat-IF dataset. For IF, we use AllenAI's RLVR-IFeval, whose prompts contain programmatically verifiable constraints. Each domain contributes 7.5K training examples, resulting in a balanced mixture across all four domains.

\paragraph{Evaluation.}
We evaluate mathematical reasoning on AIME 2024, AIME 2025, and AMC 2023; code generation on LiveCodeBench; IF on IFBench; and Chat on WildBench-Hard and CreativeWriting v3. For AIME, AMC, and LiveCodeBench, we report avg@8. WildBench-Hard and CreativeWriting v3 are reported on their normalized scales. Because the training mixture contains four equally represented domains, we use a domain-macro average as the primary aggregate metric. We first average benchmarks within mathematics and Chat, then assign equal weight to the four domain scores.
This prevents a domain from receiving greater aggregate weight solely because it is represented by more benchmarks.

\subsection{Main Results}

\begin{table}[h]
\centering
\small
\caption{Main results on the Qwen3-30B-A3B mixture-of-experts backbone. Math averages AIME24, AIME25, and AMC23 avg@8; Code uses LiveCodeBench avg@8; IF uses IFBench strict-prompt accuracy; Chat averages normalized WildBench-Hard and CreativeWriting v3 scores. Domain Avg. is the unweighted mean of the four domain scores. Higher is better. Best and second-best values are bolded and underlined.}
\label{tab:moe-results}
\setlength{\tabcolsep}{5.0pt}
\renewcommand{\arraystretch}{1.08}
\begin{tabular*}{\textwidth}{@{\extracolsep{\fill}}lccccc@{}}
\toprule
Method & Math & Code & IF & Chat & \shortstack{Domain\\Avg.} \\
\midrule
\emph{Base model} & 0.5312 & 0.3310 & \bestscore{0.2100} & 0.6392 & 0.4279 \\
\midrule
GRPO & 0.5632 & 0.2603 & 0.2066 & 0.6900 & 0.4300 \\
GRPO$+$KL$_{0.001}$ & 0.5538 & 0.2540 & 0.1900 & \bestscore{0.7271} & 0.4312 \\
MGS & \secondscore{0.5757} & \secondscore{0.3858} & 0.2000 & 0.6625 & \secondscore{0.4560} \\
CGPO & 0.5670 & 0.3835 & \secondscore{0.2067} & 0.6541 & 0.4528 \\
\midrule
\method\ (ours) & \bestscore{0.5822} & \bestscore{0.4529} & 0.1933 & \secondscore{0.7003} & \bestscore{0.4822} \\
\bottomrule
\end{tabular*}
\end{table}

\paragraph{Mixture-of-experts backbone.}
Table~\ref{tab:moe-results} shows that MGS and CGPO achieve domain-macro averages of 0.4560 and 0.4528 on Qwen3-30B-A3B, respectively, substantially exceeding standard GRPO at 0.4300. \method\ further reaches 0.4822, improving over MGS, the strongest baseline, by 0.0262 points or 5.7\%. \method\ attains the best scores in mathematics and code and ranks second in Chat, while its IF score remains below the pre-RL base model. These results demonstrate that cross-step control improves over both first-order gradient surgery and curvature-guided optimization on a sparsely activated mixture-of-experts backbone.

\begin{table}[h]
\centering
\small
\caption{Main results on Qwen3-8B-Base. Domain definitions and averaging follow Table~\ref{tab:moe-results}. Higher is better. Best and second-best values are bolded and underlined.}
\label{tab:main-results}
\setlength{\tabcolsep}{5.0pt}
\renewcommand{\arraystretch}{1.08}
\begin{tabular*}{\textwidth}{@{\extracolsep{\fill}}lccccc@{}}
\toprule
Method & Math & Code & IF & Chat & \shortstack{Domain\\Avg.} \\
\midrule
\emph{Base model} & 0.1316 & 0.1751 & 0.1433 & 0.4466 & 0.2241 \\
\midrule
GRPO & 0.2906 & 0.2119 & 0.2733 & 0.6284 & 0.3511 \\
GRPO$+$KL$_{0.001}$ & 0.2750 & \bestscore{0.2249} & \bestscore{0.3000} & \bestscore{0.6802} & \secondscore{0.3700} \\
MGS & 0.2948 & 0.0883 & 0.2833 & 0.6401 & 0.3266 \\
CGPO & \secondscore{0.3094} & 0.1828 & 0.2567 & 0.6623 & 0.3528 \\
\midrule
\method\ (ours) & \bestscore{0.3111} & \secondscore{0.2173} & \secondscore{0.2933} & \secondscore{0.6723} & \bestscore{0.3735} \\
\bottomrule
\end{tabular*}
\end{table}

\paragraph{Dense backbone.}
On Qwen3-8B-Base, Table~\ref{tab:main-results} shows that \method\ achieves the highest domain-macro average of 0.3735. This corresponds to an improvement of 0.0224 points (6.4\%) over standard GRPO and 0.0035 points (0.9\%) over GRPO$+$KL$_{0.001}$, the strongest compared baseline. \method\ achieves the highest mathematics score and ranks second in code, IF, and Chat. GRPO$+$KL$_{0.001}$ remains slightly stronger in these three domains, while its lower mathematics score results in a smaller equal-domain average. The aggregate result therefore reflects a favorable cross-domain trade-off despite the absence of uniform per-domain dominance.

\begin{table}[h]
\centering
\small
\caption{Ablation results on Qwen3-8B-Base. \method\ without history residual retains the one-step-one-lead focus structure while removing the checkpoint-history correction; the remaining variants ablate focus weighting, drift-based ranking, or token-level allocation.}
\label{tab:ablation}
\setlength{\tabcolsep}{5.0pt}
\renewcommand{\arraystretch}{1.08}
\begin{tabular*}{\textwidth}{@{\extracolsep{\fill}}lccccc@{}}
\toprule
Method & Math & Code & IF & Chat & \shortstack{Domain\\Avg.} \\
\midrule
\method\ without history residual & 0.2996 & \secondscore{0.2114} & \secondscore{0.2966} & 0.6324 & 0.3600 \\
Equal task weights & 0.0615 & 0.0202 & 0.2467 & 0.3978 & 0.1815 \\
Response-level residual & \bestscore{0.3202} & 0.2021 & \bestscore{0.3000} & \secondscore{0.6338} & \secondscore{0.3640} \\
Random residual & 0.0698 & 0.0390 & 0.2400 & 0.3866 & 0.1838 \\
\midrule
\method\ (ours) & \secondscore{0.3111} & \bestscore{0.2173} & 0.2933 & \bestscore{0.6723} & \bestscore{0.3735} \\
\bottomrule
\end{tabular*}
\end{table}

\subsection{Ablation Analysis}
The matched \method\ without history residual variant retains the same one-step-one-lead focus schedule and asymmetric focus--non-focus base weighting while setting the checkpoint-history residual to zero. It reaches a domain-macro average of 0.3600, above standard GRPO at 0.3511 (Table~\ref{tab:main-results}) but 0.0135 below the full method at 0.3735. This comparison clarifies the role of the one-step-one-lead structure in the full design: focus scheduling and asymmetric weighting establish a structured cross-step trajectory with a designated focus domain, while the checkpoint-history residual uses the preceding update to counter sequential rebound. The full method improves over this structure-only baseline by 3.8\%, indicating that focus scheduling alone does not account for the observed gain.

The remaining ablations isolate how this cross-step correction is implemented. Equal task weights keeps the focus-restricted residual but removes asymmetric base weighting, reducing the domain-macro average to 0.1815 ($-51.4\%$); Random residual keeps the correction pathway but replaces drift-based token ranking with random weights, yielding 0.1838 ($-50.8\%$). The large degradations indicate that the history correction depends on both the focus-domain weighting and the preceding footprint used to localize the correction. Response-level residual retains both the one-step-one-lead structure and checkpoint signal but removes token-specific allocation within each response, reaching 0.3640. The full token-level method further improves this score by 0.0095 points (2.6\%), mainly through gains in code and Chat. Together, the ablations support the intended decomposition of \method: one-step-one-lead organizes the cross-step trajectory, checkpoint drift localizes rebound risk, and token-level allocation applies the correction at the granularity where that risk is observed.

\begin{figure}[t]
  \centering
  \includegraphics[width=0.9\linewidth]{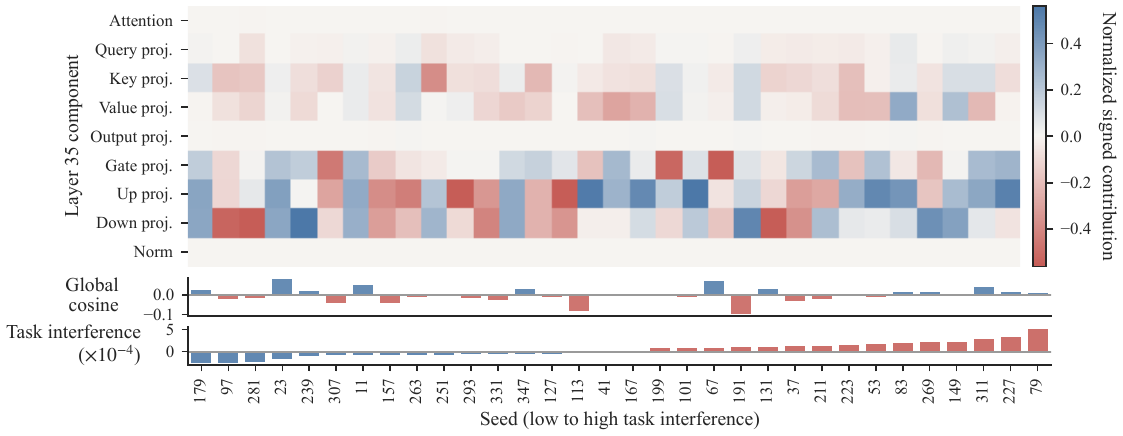}
  \caption{\textbf{Layer-35 signed gradient contributions expose cancellation hidden by the global cosine.} Columns are seeds ordered by observed task interference; opposing module contributions can cancel under global aggregation.}
  \label{fig:module-cancellation}
  \vspace{-1.0em}
\end{figure}

\begin{figure}[t]
  \centering
  \includegraphics[width=0.9\linewidth]{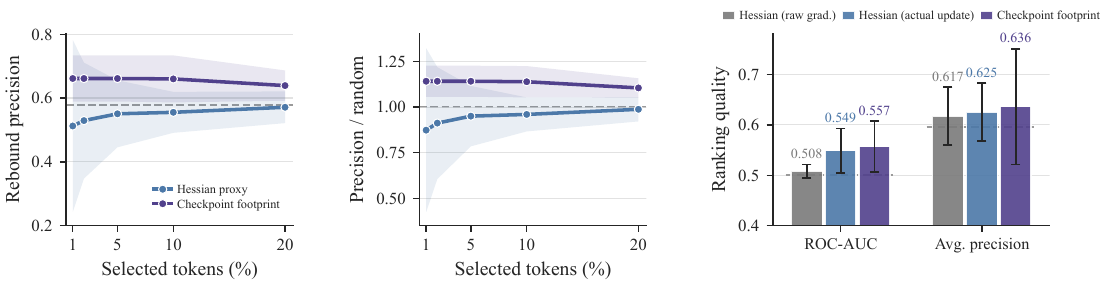}
\caption{Checkpoint history provides a stronger token-level rebound signal than Hessian-based proxies. Checkpoint-footprint ranking improves precision and enrichment across candidate fractions (left) and remains superior in ROC-AUC and average precision even when the Hessian proxy uses the realized optimizer update (right).}
  \label{fig:mechanism-evidence}
\end{figure}

\subsection{Mechanism Analysis}
\label{sec:mechanism}
Section~\ref{sec:theory} characterizes cross-step backtracking using two consecutive token footprints, whereas the online controller must act before the current update is completed and therefore observes only the preceding footprint $\bm u$. We therefore evaluate whether $\bm u$ alone contains sufficient information to identify tokens likely to rebound under the next update. As shown in Figure~\ref{fig:curvature-cost-quality}, within the negative-drift candidate set, checkpoint-footprint ranking achieves $0.567$ ROC-AUC and $0.627$ average precision, compared with $0.498$ and $0.579$ for the Hessian proxy, while improving top-$5\%$ precision from $0.551$ to $0.661$; the non-token-resolved global reference yields the expected ROC-AUC of $0.500$. These results indicate that preceding negative drift provides an informative token-level ranking of future rebound risk. Figure~\ref{fig:module-cancellation} further explains why this information can be obscured by global gradient aggregation: large opposing module-level contributions, particularly in MLP projections, cancel to produce near-zero global cosine values despite substantially different levels of task interference. Figure~\ref{fig:mechanism-evidence} further evaluates the robustness of the token-level ranking. Checkpoint-footprint ranking remains stronger across selected candidate fractions from $1\%$ to $20\%$, with the largest advantage among the highest-ranked candidates, supporting the drift-ranked token allocation in Eq.~(\ref{eq:m_quant}). On the three-seed control subset, replacing the raw gradient with the realized optimizer displacement improves the Hessian proxy's ROC-AUC from $0.508$ to $0.549$, but checkpoint ranking remains higher at $0.557$. Together, these results show that the preceding checkpoint footprint provides a direct pre-update signal for identifying rebound-prone tokens and allocating \method's correction strength without explicit Hessian estimation.

\section{Conclusion}

Multi-domain RL can suffer sequential interference even when same-point gradients appear weakly conflicting. We introduced a cross-step view in which adjacent checkpoint footprints expose the mixed output-space interaction between consecutive updates and give harmful reversals a direct interpretation as cross-step output backtracking. Building on this view, \method\ designates a focus domain and adds a drift-ranked, scale-normalized token-level history residual to the standard GRPO coefficient, requiring neither auxiliary supervision nor explicit higher-order differentiation. Our analysis characterizes when this correction contracts the targeted backtracking component. Controlled studies show that cross-step backtracking is more strongly associated with subsequent task damage than same-point gradient diagnostics, while the preceding checkpoint footprint identifies future token rebound more effectively than Hessian-based proxies. On Qwen3-30B-A3B and Qwen3-8B-Base, \method\ achieves domain-macro averages of 0.4822 and 0.3735, with relative gains of 5.7\% and 0.9\% over the strongest compared baselines, respectively. Ablations further verify the importance of focus--non-focus base weighting, drift-guided localization, and token-level allocation. Together, these results support cross-step output-space regulation as a practical first-order approach to mitigating interference in multi-domain policy optimization.

\clearpage
\bibliography{iclr2026_conference}

@article{yang2026local,
  title={A Local Perturbation Theory for Cross-Domain Interference and Recovery in Multi-Domain RL},
  author={Yang, Lei and Ding, Siyu and Xiong, Deyi},
  journal={arXiv preprint arXiv:2606.02398},
  year={2026}
}

@article{liu2021conflict,
  title={Conflict-averse gradient descent for multi-task learning},
  author={Liu, Bo and Liu, Xingchao and Jin, Xiaojie and Stone, Peter and Liu, Qiang},
  journal={Advances in neural information processing systems},
  volume={34},
  pages={18878--18890},
  year={2021}
}

@inproceedings{liang2026boosting,
  title={Boosting multi-domain reasoning of llms via curvature-guided policy optimization},
  author={Liang, Xize and Yang, Lin and Wang, Jie and Liu, Rui and Lu, Yang and Zeng, Jinliang and Chen, Hanzhu and Li, Dong and Hao, Jianye},
  booktitle={The Fourteenth International Conference on Learning Representations},
  year={2026}
}

@article{amari1998natural,
  title={Natural gradient works efficiently in learning},
  author={Amari, Shun-Ichi},
  journal={Neural computation},
  volume={10},
  number={2},
  pages={251--276},
  year={1998},
  publisher={MIT Press}
}

@article{martens2020new,
  title={New insights and perspectives on the natural gradient method},
  author={Martens, James},
  journal={Journal of Machine Learning Research},
  volume={21},
  number={146},
  pages={1--76},
  year={2020}
}

@article{kunstner2019limitations,
  title={Limitations of the empirical fisher approximation for natural gradient descent},
  author={Kunstner, Frederik and Hennig, Philipp and Balles, Lukas},
  journal={Advances in neural information processing systems},
  volume={32},
  year={2019}
}

@article{kakade2001natural,
  title={A natural policy gradient},
  author={Kakade, Sham M},
  journal={Advances in neural information processing systems},
  volume={14},
  year={2001}
}

@inproceedings{schulman2015trust,
  title={Trust region policy optimization},
  author={Schulman, John and Levine, Sergey and Abbeel, Pieter and Jordan, Michael and Moritz, Philipp},
  booktitle={International conference on machine learning},
  pages={1889--1897},
  year={2015},
  organization={PMLR}
}

@book{nesterov2013introductory,
  title={Introductory lectures on convex optimization: A basic course},
  author={Nesterov, Yurii},
  volume={87},
  year={2013},
  publisher={Springer Science \& Business Media}
}

@book{nocedal2006numerical,
  title={Numerical optimization},
  author={Nocedal, Jorge and Wright, Stephen J},
  year={2006},
  publisher={Springer}
}

@article{sutton1999policy,
  title={Policy gradient methods for reinforcement learning with function approximation},
  author={Sutton, Richard S and McAllester, David and Singh, Satinder and Mansour, Yishay},
  journal={Advances in neural information processing systems},
  volume={12},
  year={1999}
}

@article{schulman2017proximal,
  title={Proximal policy optimization algorithms},
  author={Schulman, John and Wolski, Filip and Dhariwal, Prafulla and Radford, Alec and Klimov, Oleg},
  journal={arXiv preprint arXiv:1707.06347},
  year={2017}
}

@article{jacot2018neural,
  title={Neural tangent kernel: Convergence and generalization in neural networks},
  author={Jacot, Arthur and Gabriel, Franck and Hongler, Cl{\'e}ment},
  journal={Advances in neural information processing systems},
  volume={31},
  year={2018}
}

@article{hoeffding1963probability,
  title={Probability inequalities for sums of bounded random variables},
  author={Hoeffding, Wassily},
  journal={Journal of the American statistical association},
  volume={58},
  number={301},
  pages={13--30},
  year={1963},
  publisher={Taylor \& Francis}
}

@article{basant2025nvidia,
  title={Nvidia nemotron nano 2: An accurate and efficient hybrid mamba-transformer reasoning model},
  author={Basant, Aarti and Khairnar, Abhijit and Paithankar, Abhijit and Khattar, Abhinav and Renduchintala, Adithya and Malte, Aditya and Bercovich, Akhiad and Hazare, Akshay and Rico, Alejandra and Ficek, Aleksander and others},
  journal={arXiv preprint arXiv:2508.14444},
  year={2025}
}

@article{cai2026advancing,
  title={Advancing General-Purpose Reasoning Models with Modular Gradient Surgery},
  author={Cai, Min and Liang, Yu and Wang, Longzheng and Wang, Yan and Zhang, Yueyang and Xia, Long and Sun, Zhiyuan and Ye, Xi and Shi, Daiting},
  journal={arXiv preprint arXiv:2602.02301},
  year={2026}
}

@article{guo2025deepseek,
  title={Deepseek-r1: Incentivizing reasoning capability in llms via reinforcement learning},
  author={Guo, Daya and Yang, Dejian and Zhang, Haowei and Song, Junxiao and Zhang, Ruoyu and Xu, Runxin and Zhu, Qihao and Ma, Shirong and Wang, Peiyi and Bi, Xiao and others},
  journal={arXiv preprint arXiv:2501.12948},
  year={2025}
}

@article{sener2018multi,
  title={Multi-task learning as multi-objective optimization},
  author={Sener, Ozan and Koltun, Vladlen},
  journal={Advances in neural information processing systems},
  volume={31},
  year={2018}
}

@article{shao2024deepseekmath,
  title={Deepseekmath: Pushing the limits of mathematical reasoning in open language models},
  author={Shao, Zhihong and Wang, Peiyi and Zhu, Qihao and Xu, Runxin and Song, Junxiao and Bi, Xiao and Zhang, Haowei and Zhang, Mingchuan and Li, YK and Wu, Yang and others},
  journal={arXiv preprint arXiv:2402.03300},
  year={2024}
}

@inproceedings{wu2026imbalanced,
  title={Imbalanced gradients in rl post-training of multi-task llms},
  author={Wu, Runzhe and Samanta, Ankur and Jain, Ayush and Fujimoto, Scott and Kwon, Jeongyeol and Kretzu, Ben and Yu, Youliang and Hassani, Kaveh and Vidolov, Boris and Efroni, Yonathan},
  booktitle={Findings of the Association for Computational Linguistics: EACL 2026},
  pages={3137--3150},
  year={2026}
}

@inproceedings{chen2018gradnorm,
  title={Gradnorm: Gradient normalization for adaptive loss balancing in deep multitask networks},
  author={Chen, Zhao and Badrinarayanan, Vijay and Lee, Chen-Yu and Rabinovich, Andrew},
  booktitle={International conference on machine learning},
  pages={794--803},
  year={2018},
  organization={PMLR}
}

@inproceedings{gong2024coba,
  title={Coba: Convergence balancer for multitask finetuning of large language models},
  author={Gong, Zi and Yu, Hang and Liao, Cong and Liu, Bingchang and Chen, Chaoyu and Li, Jianguo},
  booktitle={Proceedings of the 2024 Conference on Empirical Methods in Natural Language Processing},
  pages={8063--8077},
  year={2024}
}

@article{navon2022multi,
  title={Multi-task learning as a bargaining game},
  author={Navon, Aviv and Shamsian, Aviv and Achituve, Idan and Maron, Haggai and Kawaguchi, Kenji and Chechik, Gal and Fetaya, Ethan},
  journal={arXiv preprint arXiv:2202.01017},
  year={2022}
}

@article{liu2023famo,
  title={Famo: Fast adaptive multitask optimization},
  author={Liu, Bo and Feng, Yihao and Stone, Peter and Liu, Qiang},
  journal={Advances in Neural Information Processing Systems},
  volume={36},
  pages={57226--57243},
  year={2023}
}

@article{ramesh2026multi,
  title={Multi-Task GRPO: Reliable LLM Reasoning Across Tasks},
  author={Ramesh, Shyam Sundhar and Ji, Xiaotong and Zimmer, Matthieu and Yoon, Sangwoong and Wang, Zhiyong and Ammar, Haitham Bou and Lucchi, Aurelien and Bogunovic, Ilija},
  journal={arXiv preprint arXiv:2602.05547},
  year={2026}
}

@article{yang2026transferability,
  title={Transferability for General Reasoning: An Automated Curriculum for Multi-Domain RLVR},
  author={Yang, Yongjin and Liu, Jiarui and He, Yinghui and Zhang, Lezhen and Sch{\"o}lkopf, Bernhard and Jin, Zhijing},
  journal={arXiv preprint arXiv:2606.25178},
  year={2026}
}

@inproceedings{koh2017understanding,
  title={Understanding black-box predictions via influence functions},
  author={Koh, Pang Wei and Liang, Percy},
  booktitle={International conference on machine learning},
  pages={1885--1894},
  year={2017},
  organization={PMLR}
}

@article{pruthi2020estimating,
  title={Estimating training data influence by tracing gradient descent},
  author={Pruthi, Garima and Liu, Frederick and Kale, Satyen and Sundararajan, Mukund},
  journal={Advances in Neural Information Processing Systems},
  volume={33},
  pages={19920--19930},
  year={2020}
}

@article{cheng2026revisiting,
  title={Revisiting reinforcement learning for llm reasoning from a cross-domain perspective},
  author={Cheng, Jorge Zhoujun and Hao, Shibo and Liu, Tianyang and Zhou, Fan and Xie, Yutao and Yao, Feng and Bian, Yuexin and Dey, Nilabjo and Zhuang, Yonghao and Zha, Yuheng and others},
  journal={Advances in Neural Information Processing Systems},
  volume={38},
  year={2026}
}

@article{li2025can,
  title={Can one domain help others? a data-centric study on multi-domain reasoning via reinforcement learning},
  author={Li, Yu and Pan, Zhuoshi and Lin, Honglin and Sun, Mengyuan and He, Conghui and Wu, Lijun},
  journal={arXiv preprint arXiv:2507.17512},
  year={2025}
}

@article{yu2026dapo,
  title={Dapo: An open-source llm reinforcement learning system at scale},
  author={Yu, Qiying and Zhang, Zheng and Zhu, Ruofei and Yuan, Yufeng and Zuo, Xiaochen and Yue, Yu and Dai, Weinan and Fan, Tiantian and Liu, Gaohong and Liu, Lingjun and others},
  journal={Advances in Neural Information Processing Systems},
  volume={38},
  pages={113222--113244},
  year={2026}
}

@article{yang2026domains,
  title={When Domains Interact: Asymmetric and Order-Sensitive Cross-Domain Effects in Reinforcement Learning for Reasoning},
  author={Yang, Wang and Wang, Shouren and Song, Chaoda and Ma, Chuang and Li, Xinpeng and Wang, Nengbo and Zhou, Kaixiong and Chaudhary, Vipin and Han, Xiaotian},
  journal={arXiv preprint arXiv:2602.01365},
  year={2026}
}

@article{yu2020gradient,
  title={Gradient surgery for multi-task learning},
  author={Yu, Tianhe and Kumar, Saurabh and Gupta, Abhishek and Levine, Sergey and Hausman, Karol and Finn, Chelsea},
  journal={Advances in neural information processing systems},
  volume={33},
  pages={5824--5836},
  year={2020}
}

@misc{lin2026resrlboostingllmreasoning,
      title={ResRL: Boosting LLM Reasoning via Negative Sample Projection Residual Reinforcement Learning}, 
      author={Zihan Lin and Xiaohan Wang and Jie Cao and Jiajun Chai and Li Wang and Xiaodong Lu and Wei Lin and Ran He and Guojun Yin},
      year={2026},
      eprint={2605.00380},
      archivePrefix={arXiv},
      primaryClass={cs.LG},
      url={https://arxiv.org/abs/2605.00380}, 
}

@inproceedings{ lin2026rest, title={ResT: Reshaping Token-Level Policy Gradients for Tool-Use Large Language Models}, author={Zihan Lin and Xiaohan Wang and Jie Cao and Jiajun Chai and Guojun Yin and Wei Lin and Ran He}, booktitle={The Fourteenth International Conference on Learning Representations}, year={2026}, url={https://openreview.net/forum?id=gNZlaKRWki} }

@article{liu2026cdrrm,   title={CDRRM: Contrast-Driven Rubric Generation for Reliable and Interpretable Reward Modeling},   author={Liu, Dengcan and Yang, Fengkai and Wang, Xiaohan and Yan, Shurui and Chai, Jiajun and Li, Jiahao and Ban, Yikun and Mao, Zhendong and Lin, Wei and Yin, Guojun},   journal={arXiv preprint arXiv:2603.08035},   year={2026} }
\bibliographystyle{plainnat}

\clearpage
\appendix

\section{Controlled Diagnostic Study Protocols}
\label{app:diagnostic_protocols}

This section provides the complete protocols for the two controlled studies used to motivate and analyze \method. Both studies instantiate adjacent updates explicitly, preserve the optimizer state across the two steps, and evaluate every diagnostic against quantities measured from the realized checkpoint trajectory. The first study evaluates which replicate-level signal best explains downstream cross-domain interference. The second evaluates which token-level signal best identifies tokens whose preceding output movement is reversed by the next update.

\subsection{Common model, data, and one-step optimization}
\label{app:diagnostic_common}

\paragraph{Model and data.}
We use Qwen3-4B as the policy in both studies. Math prompts are sampled from the math portion of the same filtered NVIDIA Nemotron Post-Training Dataset v2 mixture used for the main experiments, and chat prompts are sampled from the MGS-provided WildChat-IF subset. Math generations are scored by the task verifier and chat generations are scored by Skywork-Reward-Llama-3.1-8B-v0.2. For each prompt, we sample four responses with temperature $1.0$ and nucleus threshold $p=0.95$. Prompts are truncated to at most $1{,}024$ tokens and responses to at most $512$ tokens. We disable explicit thinking-mode prompting so that all diagnostics are computed under the same response format.

\paragraph{Advantages and updates.}
Within each four-response group, rewards are standardized to produce the GRPO trajectory advantage. If every response in a group receives the same verifier reward, we use a unit coefficient for that group, yielding a deterministic nonzero one-step update while preserving the sampled response distribution. Chat groups use reward-model scores and have nonzero within-group variation in the analyzed runs. Each replicate performs two consecutive AdamW steps with learning rate $10^{-6}$, weight decay $0.1$, $(\beta_1,\beta_2)=(0.9,0.999)$, numerical constant $10^{-8}$, and gradient-norm clipping at $1.0$. Computation uses bfloat16, and log probabilities are accumulated over response tokens with padding coordinates masked as appropriate.

\paragraph{Controlled output-facing subspace.}
The repeated gradient, optimizer, and finite-difference evaluations are carried out in an output-facing parameter subspace consisting of the final transformer block and the language-model head. This subspace contains $100{,}930{,}816$ parameters, or $2.51\%$ of Qwen3-4B, and is shared by the compared diagnostics in the primary studies. The coverage control in Appendix~\ref{app:preexp2_controls} expands this subspace to the final four and final eight transformer blocks while retaining the language-model head. This design permits exact replicate-level replay of gradients, optimizer displacements, and token log probabilities while keeping the policy, sampled trajectories, rewards, and evaluation targets matched.

\subsection{Study I: same-point diagnostics and realized task interference}
\label{app:preexp1_protocol}

\paragraph{Replicate construction.}
We construct $32$ independent replicates from a common initial checkpoint $\bm\theta_0$. Randomness across replicates comes from prompt selection and response sampling; the initial policy and all optimization hyperparameters are fixed. Each replicate uses two math update prompts and two chat update prompts, with four responses per prompt. In addition, it samples two math prompts for the token-motion probe and two disjoint math prompts for the damage probe. The update, motion-probe, and damage-probe prompt sets are mutually disjoint within a replicate.

Starting from $\bm\theta_0$, we first apply the math update and then the chat update:
\begin{equation}
\label{eq:preexp1_triplet}
\bm\theta_0
\xrightarrow{\;\text{math update}\;}
\bm\theta_1
\xrightarrow{\;\text{chat update}\;}
\bm\theta_2.
\end{equation}
Math responses and both math probe sets are sampled at $\bm\theta_0$. After the math step, chat responses are sampled from $\bm\theta_1$, scored by the reward model, and used for the $\bm\theta_1\!\rightarrow\!\bm\theta_2$ update. To obtain a same-point gradient comparison without changing the realized chat trajectories, the fixed $\bm\theta_1$ chat responses, rewards, and advantages are replayed at $\bm\theta_0$ when computing the chat gradient. The same AdamW optimizer instance is retained across the math and chat steps, matching the stateful sequential-update dynamics.

\paragraph{Independent interference target.}
Let $\mathcal P_{\mathrm{dmg}}$ denote the held-out math damage probe. We measure the effect of the chat update by the change in fixed-response token negative log likelihood:
\begin{equation}
\label{eq:preexp1_damage}
\mathrm{Damage}_{\mathrm{math}\leftarrow\mathrm{chat}}
:=
\mathcal L_{\mathrm{NLL}}(\bm\theta_2;\mathcal P_{\mathrm{dmg}})
-
\mathcal L_{\mathrm{NLL}}(\bm\theta_1;\mathcal P_{\mathrm{dmg}}).
\end{equation}
A positive value indicates that the chat step increases math-probe NLL. Because $\mathcal P_{\mathrm{dmg}}$ is disjoint from the update and motion probes, the dependent variable is not reused in any diagnostic score.

\paragraph{Same-point global and module diagnostics.}
Using the fixed update batches, define the descent directions at $\bm\theta_0$ as
\begin{equation}
\label{eq:preexp1_gradients}
\bm v_{\mathrm m}:=-\nabla\mathcal L_{\mathrm m}(\bm\theta_0),
\qquad
\bm v_{\mathrm c}:=-\nabla\mathcal L_{\mathrm c}(\bm\theta_0).
\end{equation}
The global first-order diagnostic is
\begin{equation}
\label{eq:preexp1_global_score}
s_{\mathrm{global}}
=
\left[
-
\frac{\langle\bm v_{\mathrm m},\bm v_{\mathrm c}\rangle}
{\lVert\bm v_{\mathrm m}\rVert_2\lVert\bm v_{\mathrm c}\rVert_2}
\right]_+.
\end{equation}
For the module diagnostic, partition the output-facing subspace into attention projections, MLP projections, normalization parameters, and the language-model head. If $\bm v_{\mathrm m}^{(g)}$ and $\bm v_{\mathrm c}^{(g)}$ are the directions in group $g$, define
\begin{equation}
\label{eq:preexp1_module_score}
s_{\mathrm{module}}
=
\frac{\sum_g[-\langle\bm v_{\mathrm m}^{(g)},\bm v_{\mathrm c}^{(g)}\rangle]_+}
{\sum_g|\langle\bm v_{\mathrm m}^{(g)},\bm v_{\mathrm c}^{(g)}\rangle|}.
\end{equation}
This normalized negative dot-product mass retains opposing module contributions that cancel in the aggregate inner product.

\paragraph{Cross-step backtracking diagnostic.}
On the disjoint math motion probe $\mathcal P_{\mathrm{mot}}$, we evaluate the same realized response tokens at all three checkpoints. For token $i$, let
\begin{equation}
\label{eq:preexp1_motion}
u_i
=
\log\pi_{\bm\theta_1}(a_i\mid x_i)
-
\log\pi_{\bm\theta_0}(a_i\mid x_i),
\qquad
d_i
=
\log\pi_{\bm\theta_2}(a_i\mid x_i)
-
\log\pi_{\bm\theta_1}(a_i\mid x_i).
\end{equation}
The backtracking mass used in Figure~\ref{fig:first-order-diagnostics} is
\begin{equation}
\label{eq:preexp1_reversal_mass}
s_{\mathrm{rev}}
=
\frac1{|\mathcal R_{\mathrm{mot}}|}
\sum_{i\in\mathcal R_{\mathrm{mot}}}
\mathbf 1\{u_i<0\}\,[-u_i d_i]_+
=
\frac1{|\mathcal R_{\mathrm{mot}}|}
\sum_{i\in\mathcal R_{\mathrm{mot}}}
(-u_i)[d_i]_+\,\mathbf 1\{u_i<0\},
\end{equation}
where $\mathcal R_{\mathrm{mot}}$ contains response-token coordinates. The diagnostic therefore records the mass of positive rebound following a negative preceding movement, using the realized optimizer steps rather than a virtual parameter perturbation.

\paragraph{Statistical analysis.}
Each replicate contributes one value of $\mathrm{Damage}_{\mathrm{math}\leftarrow\mathrm{chat}}$ and one value for each diagnostic. We use Spearman rank correlation as the primary association statistic and display a Theil--Sen trend in the scatter plots. Confidence intervals are obtained with $50{,}000$ paired replicate bootstrap resamples. The same resampled replicate indices are used when comparing two diagnostics, so the reported correlation gains are paired differences rather than differences between independently estimated intervals. We additionally recompute each correlation after removing every replicate once. The resulting leave-one-replicate-out ranges are reported in Figure~\ref{fig:leave-one-seed-out}.

\paragraph{Observed diagnostic ordering.}
The global and module diagnostics have correlations of $-0.19$ and $-0.22$ with downstream interference, whereas backtracking mass reaches $\rho_s=0.40$ with a 95\% bootstrap interval of $[0.05,0.69]$. Its paired correlation gains are $+0.59$ over the global diagnostic and $+0.63$ over the module diagnostic. The sign and ordering are preserved under every leave-one-replicate-out recomputation. Figure~\ref{fig:module-cancellation} further decomposes the same replicates and shows how positive and negative module contributions disappear under global aggregation.

\subsection{Study II: token-level rebound identification}
\label{app:preexp2_protocol}

\paragraph{Causal triplets and token table.}
The second study uses eight independent seeds and the same Qwen3-4B one-step configuration. Each seed samples two math prompts and two chat prompts with four responses per prompt. The math step produces $\bm\theta_1$ from $\bm\theta_0$. Chat responses are then sampled from $\bm\theta_1$, scored by the reward model, and used both as the token probe and as the update batch that produces $\bm\theta_2$. Therefore, $u_i$ and $d_i$ in Eq.~(\ref{eq:preexp1_motion}) are evaluated on the exact token coordinates responsible for the second update.

We retain chat response tokens with nonzero trajectory-level advantage:
\begin{equation}
\label{eq:preexp2_eligible}
\mathcal E
=
\{i:\ i\text{ is a response token},\  A_i\neq0\}.
\end{equation}
The checkpoint-history candidate set and realized rebound label are
\begin{equation}
\label{eq:preexp2_target}
\mathcal C
=
\{i\in\mathcal E:u_i<0\},
\qquad
y_i
=
\mathbf 1\{u_i<0,\ d_i>0\},
\qquad
q_i
=
(-u_i)[d_i]_+.
\end{equation}
The paper-facing comparison is conditioned on $\mathcal C$: it asks which negative-drift tokens are most likely to rebound under the subsequent unregulated update. Across the eight seeds, $22.6\%$ of eligible tokens enter $\mathcal C$, and the mean rebound prevalence within $\mathcal C$ is $57.8\%$.

\paragraph{Checkpoint-footprint score.}
Within $\mathcal C$, the raw checkpoint score is $[-u_i]_+$. The rank-based score used by \method\ sorts candidate drifts in ascending order and assigns
\begin{equation}
\label{eq:preexp2_quantile_score}
p_i
=
\frac{\operatorname{rank}_0(u_i)+1/2}{|\mathcal C|},
\qquad
s_i^{\mathrm{ckpt}}
=
1-
\frac{\mathrm{clip}(p_i,0.1,1)-0.1}{0.9},
\end{equation}
with score zero outside $\mathcal C$. More negative drift receives a larger score, while the strongest decile shares the maximum score. This is the same ordering and quantile profile used by the online controller before batchwise scale normalization.

\paragraph{Finite-difference Hessian proxy.}
We compare checkpoint history with a token-level curvature proxy constructed on the same fixed chat trajectories, rewards, and advantages. Let $\mathcal L_{\mathrm c}^{\mathrm{loc}}$ be the unclipped local policy-gradient surrogate on this fixed chat batch, and let
\begin{equation}
\label{eq:preexp2_directions}
\bm v_{\mathrm m}:=-\nabla\mathcal L_{\mathrm m}(\bm\theta_0),
\qquad
\bm v_{\mathrm c}:=-\nabla\mathcal L_{\mathrm c}^{\mathrm{loc}}(\bm\theta_0).
\end{equation}
For selected parameters $\bm\theta_{\mathcal S}$, choose
\begin{equation}
\label{eq:preexp2_fd_scale}
\alpha
=
\epsilon
\frac{\lVert\bm\theta_{\mathcal S}\rVert_2}
{\lVert\bm v_{\mathrm m}\rVert_2}
\end{equation}
and estimate the directional Hessian-vector product by the centered difference
\begin{equation}
\label{eq:preexp2_hvp}
\bm h
\approx
\frac{
\nabla\mathcal L_{\mathrm c}^{\mathrm{loc}}(\bm\theta_0+\alpha\bm v_{\mathrm m})
-
\nabla\mathcal L_{\mathrm c}^{\mathrm{loc}}(\bm\theta_0-\alpha\bm v_{\mathrm m})
}{2\alpha}.
\end{equation}
The main sweep uses $\epsilon\in\{10^{-5},3\!\times\!10^{-5},10^{-4},3\!\times\!10^{-4}\}$; the primary score uses the pre-specified summary entry at $\epsilon=10^{-4}$. Define
\begin{equation}
\label{eq:preexp2_virtual_points}
\bm\theta_{\mathrm m}
=
\bm\theta_0+\alpha\bm v_{\mathrm m},
\qquad
\bm\theta_{\mathrm{pred}}
=
\bm\theta_{\mathrm m}
+\alpha(\bm v_{\mathrm c}-\alpha\bm h).
\end{equation}
For probe token $i$, write $\ell_i(\bm\theta):=\log\pi_{\bm\theta}(a_i\mid x_i)$. The virtual math displacement and curvature-adjusted predicted chat displacement are
\begin{equation}
\label{eq:preexp2_virtual_motion}
\Delta_i^{\mathrm m}
=
\ell_i(\bm\theta_{\mathrm m})-\ell_i(\bm\theta_0),
\qquad
\Delta_i^{\mathrm{pred}}
=
\ell_i(\bm\theta_{\mathrm{pred}})-\ell_i(\bm\theta_{\mathrm m}),
\end{equation}
and the Hessian ranking score is
\begin{equation}
\label{eq:preexp2_hessian_score}
s_i^{\mathrm H}
=
[-\Delta_i^{\mathrm m}\Delta_i^{\mathrm{pred}}]_+.
\end{equation}
All virtual log probabilities are computed on the same tokenized probe batch. We also record the induced forward policy KL for each virtual displacement, providing an output-space check on the numerical perturbation scale.

\paragraph{Ranking metrics and uncertainty.}
For every seed, we compute receiver operating characteristic area under the curve (ROC-AUC), average precision, and precision among the top-ranked $1\%$, $2\%$, $5\%$, $10\%$, and $20\%$ of candidates. Enrichment is precision divided by the within-candidate rebound prevalence. Average precision is evaluated at unique score thresholds, and when a top-$k$ boundary intersects tied scores, we report expected precision under uniform tie breaking. We first compute each metric independently for each seed and then report the seed mean with a two-sided 95\% Student-$t$ interval. Paired gains use the per-seed difference between checkpoint-footprint and Hessian scores before aggregation.

\paragraph{Primary comparison.}
Within negative-drift candidates, checkpoint-footprint ranking obtains $0.567$ ROC-AUC, $0.627$ average precision, and $0.661$ top-$5\%$ precision, compared with $0.498$, $0.579$, and $0.551$ for the Hessian proxy. The paired gains in ROC-AUC and average precision are $0.069$ and $0.048$, respectively. Across selected candidate fractions, checkpoint ranking remains above the Hessian proxy in both precision and enrichment, with the clearest separation when only a small fraction of candidates is selected by \method.

\subsection{Hessian controls and computational measurements}
\label{app:preexp2_controls}

\paragraph{Realized-update direction.}
To separate curvature approximation from direction mismatch, we repeat the Hessian construction on seeds $\{11,23,37\}$ after replacing $\bm v_{\mathrm m}$ with the selected-parameter optimizer displacement $\bm\theta_1-\bm\theta_0$. The finite-difference procedure and token target remain unchanged. This increases Hessian ROC-AUC from $0.508$ to $0.549$ and average precision from $0.617$ to $0.625$; checkpoint footprint remains highest at $0.557$ and $0.636$, respectively.

\paragraph{Parameter-coverage sweep.}
We repeat the curvature proxy with the language-model head plus the final one, four, or eight transformer blocks, corresponding to $2.51\%$, $10.04\%$, and $20.07\%$ of model parameters. Each configuration uses the same three seeds and finite-difference scales $\{3\!\times\!10^{-5},10^{-4}\}$. Peak GPU memory is measured by the maximum CUDA allocation during the complete diagnostic run. Mean peak memory rises from $26.6$ to $33.8$ and $43.5$ GB, while Hessian ROC-AUC changes from $0.508$ to $0.511$ and $0.534$. Checkpoint-footprint ROC-AUC on the matching trajectories is $0.557$, $0.577$, and $0.583$.

\paragraph{Finite-difference scale sweep.}
Finally, we retain all eight primary seeds and recompute the finite-difference diagnostic for $\epsilon\in\{10^{-5},3\!\times\!10^{-5},10^{-4},3\!\times\!10^{-4}\}$. For each scale we record the HVP norm and the forward policy KL induced by the virtual math step, the displacement induced by the negative HVP term, and predicted next step. These measurements produce Figure~\ref{fig:hvp-epsilon-sensitivity} and quantify how the numerical probe scale changes the output-space displacement being evaluated.

\clearpage
\section{Additional Diagnostic Evidence}
\label{app:diagnostic_evidence}

\begin{figure}[t]
  \centering
  \includegraphics[width=0.9\linewidth]{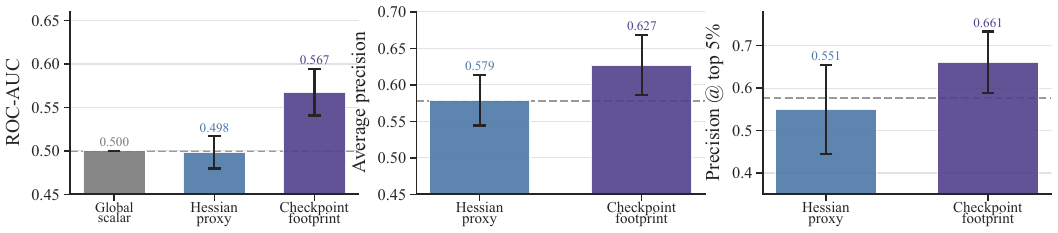}
  \caption{Rebound identification within negative-drift candidates. The checkpoint footprint outperforms the Hessian proxy in ROC-AUC, average precision, and top-5\% precision; the non-token-resolved global scalar is included as a 0.5 ROC-AUC reference. Error bars show 95\% confidence intervals.}
  \label{fig:curvature-cost-quality}
\end{figure}

This section expands the controlled analyses in Appendix~\ref{app:diagnostic_protocols} and clarifies what each diagnostic reveals about cross-step interference. Study~I evaluates \emph{realized} two-step backtracking after both updates have occurred and asks whether it is associated with downstream task damage; this is a post-update diagnostic question rather than the pre-update token-ranking problem used by the online controller. Study~II examines the numerical behavior of the finite-difference Hessian proxy used as a token-level comparator. Together, the analyses separate three conceptually distinct questions: whether same-point gradients order realized damage, whether the observed diagnostic ordering is stable across replicates, and whether explicit curvature estimation provides a numerically stable route to token-level interaction information.

\begin{figure}[h]
  \centering
  \includegraphics[width=0.95\linewidth]{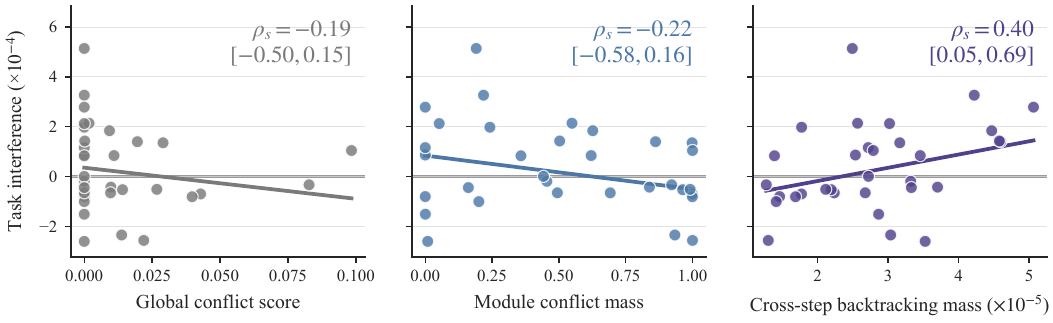}
  \caption{Replicate-level association between interference diagnostics and realized task damage. Same-point global and module scores do not positively order the subsequent damage, whereas cross-step backtracking mass shows a positive association ($\rho_s=0.40$, 95\% bootstrap CI $[0.05,0.69]$). The backtracking and damage measurements use disjoint probes.}
  \label{fig:first-order-diagnostics}
\end{figure}

\paragraph{Cross-step backtracking is more closely associated with realized task damage.}
Figure~\ref{fig:first-order-diagnostics} compares the three replicate-level diagnostics from Study~I against the held-out math damage measured after the subsequent chat update. The same-point global conflict score has Spearman correlation $\rho_s=-0.19$ with damage, while the module-level conflict mass gives $\rho_s=-0.22$. These negative estimates should not be interpreted as evidence that gradient conflict is beneficial; rather, within this controlled setting, neither same-point score provides a useful positive ordering of which subsequent updates damage the held-out math probe more strongly. This remains true even for the module diagnostic, which preserves opposing module contributions that disappear in the global inner product. By contrast, the realized cross-step backtracking mass reaches $\rho_s=0.40$ with a 95\% bootstrap CI of $[0.05,0.69]$: replicates in which the second update more strongly reverses the output movement of the first tend to exhibit larger downstream damage. Importantly, the backtracking score is computed on the motion probe, whereas damage is measured on a disjoint held-out probe (Appendix~\ref{app:preexp1_protocol}); the association therefore does not arise from evaluating the same token set twice. The result supports the central diagnostic claim of the paper: the interaction between two \emph{realized} updates in output space is more strongly associated with sequential task damage than same-checkpoint gradient geometry alone.

\begin{figure}[h]
  \centering
  \includegraphics[width=\linewidth]{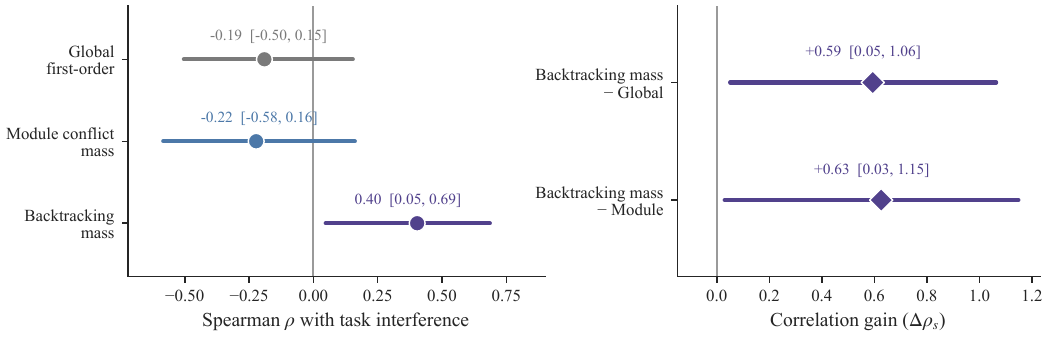}
  \caption{Uncertainty and paired correlation gains for the three Study~I diagnostics. Left: bootstrap estimates of Spearman correlation with task damage. Right: paired gains from replacing the same-point global or module diagnostic with cross-step backtracking mass. Both paired intervals remain positive.}
  \label{fig:correlation-evidence}
\end{figure}

\paragraph{Paired uncertainty favors the trajectory-based diagnostic.}
Figure~\ref{fig:correlation-evidence} makes the uncertainty comparison explicit. The left panel reports the bootstrap intervals of the three Spearman correlations: the intervals for the global and module diagnostics include zero, while the interval for backtracking mass lies on the positive side. More importantly, the right panel compares the diagnostics \emph{pairwise} using identical bootstrap resamples of the 32 replicates. Backtracking mass improves correlation by $+0.59$ over the global score (95\% CI $[0.05,1.06]$) and by $+0.63$ over the module score (95\% CI $[0.03,1.15]$). Because these are paired differences, the comparison controls for replicate-to-replicate variation rather than contrasting two independently estimated confidence intervals. Combined with the module decomposition in Figure~\ref{fig:module-cancellation}, this result also separates two phenomena: module-level decomposition can reveal cancellation hidden by a global cosine, but retaining that cancellation information alone is still insufficient to match the realized two-step backtracking signal in explaining downstream damage. The added value therefore comes from observing the trajectory interaction itself, not merely from computing a more granular same-point gradient statistic.

\begin{figure}[h]
  \centering
  \includegraphics[width=0.85\linewidth]{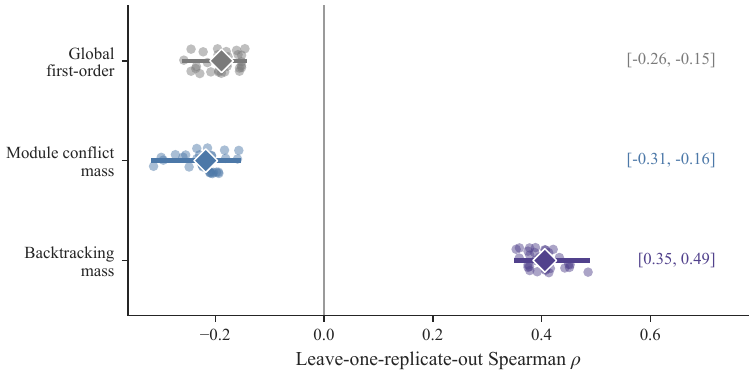}
  \caption{Leave-one-replicate stability of the Study~I correlations. Each point recomputes Spearman $\rho$ after removing one replicate; horizontal ranges summarize all 32 deletions. The sign and ordering of the diagnostics are preserved throughout.}
  \label{fig:leave-one-seed-out}
\end{figure}

\paragraph{The diagnostic ordering is not driven by a single replicate.}
The controlled study contains 32 replicates, so we additionally assess sensitivity to individual runs rather than relying on the aggregate correlation alone. Figure~\ref{fig:leave-one-seed-out} recomputes each Spearman correlation after removing one replicate at a time. Across all 32 leave-one-out subsets, the global score remains negative in $[-0.26,-0.15]$, the module conflict mass remains negative in $[-0.31,-0.16]$, and backtracking mass remains positive in $[0.35,0.49]$. Thus every deletion preserves both the sign pattern and the ordering of the three diagnostics. This analysis is not a substitute for an independent replication study, but it rules out the simpler explanation that the positive backtracking correlation, or the gap to the same-point baselines, is caused by one unusually influential replicate.

\begin{figure}[h]
  \centering
  \includegraphics[width=0.9\linewidth]{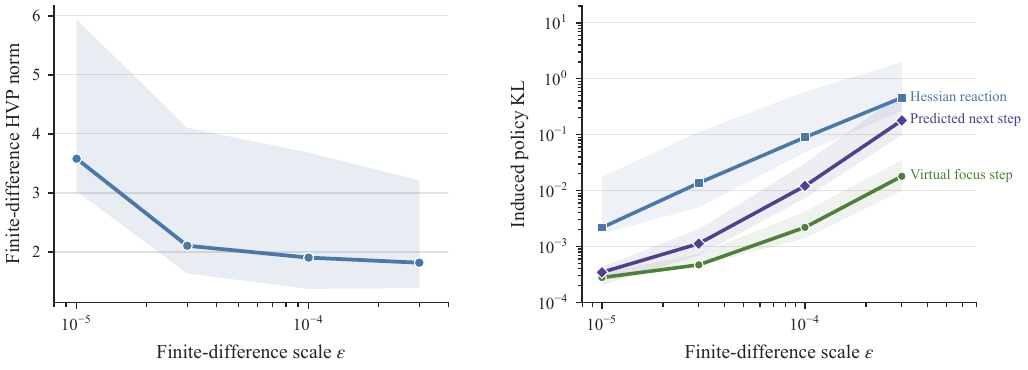}
  \caption{Sensitivity of the finite-difference Hessian proxy to the probe scale $\epsilon$. Left: the estimated HVP norm varies with $\epsilon$. Right: the induced policy KL of the virtual focus step, Hessian reaction, and predicted next step changes by orders of magnitude, showing that the probe scale also changes the output-space displacement being evaluated.}
  \label{fig:hvp-epsilon-sensitivity}
\end{figure}

\paragraph{Finite-difference curvature introduces a consequential probe-scale choice.}
Figure~\ref{fig:hvp-epsilon-sensitivity} examines the finite-difference scale $\epsilon$ used to construct the Hessian-vector-product proxy in Study~II. The left panel shows that the estimated HVP norm changes markedly at the smallest probe scale and then becomes flatter as $\epsilon$ increases. The right panel shows the corresponding output-space effect: the policy KL induced by the virtual focus step, the Hessian reaction, and the resulting predicted next step all increase by orders of magnitude over the same sweep. Hence $\epsilon$ affects not only the numerical value of the curvature estimate but also the size of the virtual policy displacement on which the token-ranking proxy is evaluated. This introduces an additional sensitivity for online use: changing the probe scale changes the effective output-space regime of the diagnostic even when the underlying checkpoint and trajectories are fixed. The checkpoint-footprint signal used by \method\ does not require this additional finite-difference probe; it is measured directly from the log-probability change between realized adjacent checkpoints. Figure~\ref{fig:hvp-epsilon-sensitivity} therefore complements the accuracy and memory controls in Appendix~\ref{app:preexp2_controls}: the limitation of the Hessian route is not only computational cost, but also the need to choose a numerical perturbation scale whose effect propagates into the policy-space quantity being ranked.

Taken together, these controls sharpen the mechanism evidence without changing the role assigned to each study in the main text. Study~I shows that realized cross-step backtracking is a stronger \emph{post-update diagnostic} of task damage than same-point gradient scores. Study~II shows that finite-difference curvature introduces an additional probe-scale choice whose effect propagates into the output-space quantity used for token ranking. The online controller instead uses the preceding checkpoint footprint, which is already available before the next update and whose rebound-ranking ability is evaluated in Section~\ref{sec:mechanism}. The results are therefore complementary: realized two-step backtracking characterizes the interaction that occurred, while the preceding footprint provides the pre-update information used to determine the intervention set.
\FloatBarrier
\clearpage

\section{Output-Space Geometry}
\label{app:output_geometry}

\subsection{Notation and local regularity}
\label{app:standing}

Fix iteration $k$ and the eligible focus coordinates $\mathcal E_k=\{(x_i,a_i)\}_{i=1}^{N}$. For an arbitrary action $a$ at context $x_i$, define
\begin{equation}
\label{eq:app_action_logprob}
\ell_{i,a}(\bm\theta):=\log\pi_{\bm\theta}(a\mid x_i),
\qquad
\score_{i,a}:=\nabla\ell_{i,a}(\bm\theta_k).
\end{equation}
For the realized action $a_i$, abbreviate
\begin{equation}
\label{eq:app_realized_logprob}
\ell_i(\bm\theta):=\ell_{i,a_i}(\bm\theta),
\qquad
\score_i:=\score_{i,a_i},
\qquad
\bm\ell_k(\bm\theta):=[\ell_1(\bm\theta),\ldots,\ell_N(\bm\theta)]^{\top}.
\end{equation}
Collect the realized scores into
\begin{equation}
\label{eq:app_score_matrix}
\Phimat_k:=[\score_1,\ldots,\score_N]\in\mathbb R^{P\times N},
\qquad
\Kmat_k:=\Phimat_k^{\top}\Phimat_k\in\mathbb R^{N\times N}.
\end{equation}
Then
\begin{equation}
\label{eq:app_fisher_definitions}
\Fmat_k=\frac1N\Phimat_k\Phimat_k^{\top},
\qquad
\Fbar_k
=\frac1N\sum_{i=1}^{N}
\mathbb E_{a\sim\pi_{\bm\theta_k}(\cdot\mid x_i)}
[\score_{i,a}\score_{i,a}^{\top}].
\end{equation}

We use the following local conditions, standard for Taylor and trust-region analyses \citep{nesterov2013introductory,nocedal2006numerical}.
\begin{enumerate}
\item The policy support is fixed in a neighborhood $\mathcal N_k$ of $\bm\theta_k$, and $\ell_{i,a}$ is three times continuously differentiable on $\mathcal N_k$.
\item There are constants $G,L,M<\infty$ such that, for every relevant $(i,a)$ and $\bm\theta\in\mathcal N_k$,
\begin{equation}
\label{eq:app_regular_bounds}
\lVert\nabla\ell_{i,a}(\bm\theta)\rVert\leq G,
\qquad
\lVert\nabla^2\ell_{i,a}(\bm\theta)\rVert_{\mathrm{op}}\leq L,
\qquad
\lVert\nabla^3\ell_{i,a}(\bm\theta)\rVert_{\mathrm{op}}\leq M.
\end{equation}
\item The line segments joining $\bm\theta_{k-1}$, $\bm\theta_k$, $\bm\theta_{k+1}^{0}$, and the locally corrected endpoint remain in $\mathcal N_k$.
\item The adjacent displacements satisfy
\begin{equation}
\label{eq:app_small_steps}
\lVert\bdelta_{k-1}\rVert\leq a\eta,
\qquad
\lVert\bdelta_k^{0}\rVert\leq b\eta,
\qquad
\lVert\bdelta_k^{r}\rVert\leq c_r\eta
\end{equation}
for constants $a,b,c_r$ independent of $\eta$.
\end{enumerate}
The checkpoint identity in Appendix~\ref{app:checkpoint_observability} is deterministic for the realized token coordinates and does not require independence across them. Sampling assumptions enter only when interpreting $\Fmat_k$ as an estimator of $\Fbar_k$.

\subsection{Relation to objective-Hessian perturbation analyses}
\label{app:objective_relation}

Prior local-perturbation work studies damage to an earlier task objective after a later domain update. Around an approximately stationary earlier-task checkpoint, Taylor expansion makes the objective-Hessian quadratic term dominant \citep{yang2026local}; curvature-guided multi-domain optimization similarly motivates explicit or approximate curvature information \citep{liang2026boosting}. These results explain why same-point gradient cosine can miss sequential interference, but their primary object is the self-quadratic sensitivity of one displacement under an objective Hessian.

Our analysis changes both the scalar function and the interaction being studied. We expand the task-conditioned output KL in Eq.~(\ref{eq:output_divergence}), whose exact Hessian at the anchor is the Fisher matrix, and then evaluate the mixed output interaction between the preceding realized displacement and the current base displacement. Consequently, Eq.~(\ref{eq:cross_duality}) is a bilinear second-order term in output geometry, not an identification of the complete objective Hessian with the Fisher. This output-space choice follows natural-gradient and trust-region policy optimization \citep{amari1998natural,kakade2001natural,schulman2015trust,martens2020new}, while making the interaction observable through checkpoint log-probabilities.

\subsection{Proof that task-conditioned output KL has Fisher Hessian}
\label{app:kl_hessian}

For one context $x_i$, write
\begin{equation}
\label{eq:app_single_kl}
D_i(\bdelta)
:=
\KL\!\left(
\pi_{\bm\theta_k}(\cdot\mid x_i)
\,\Vert\,
\pi_{\bm\theta_k+\bdelta}(\cdot\mid x_i)
\right).
\end{equation}
Let $p_i(a):=\pi_{\bm\theta_k}(a\mid x_i)$. Since the first distribution is fixed with respect to $\bdelta$,
\begin{equation}
\label{eq:app_kl_expanded}
D_i(\bdelta)
=\sum_a p_i(a)\log p_i(a)
-\sum_a p_i(a)\ell_{i,a}(\bm\theta_k+\bdelta).
\end{equation}
At $\bdelta=\bm0$, the two distributions coincide, so
\begin{equation}
\label{eq:app_kl_zero}
D_i(\bm0)=0.
\end{equation}
Differentiating Eq.~(\ref{eq:app_kl_expanded}),
\begin{equation}
\label{eq:app_kl_gradient}
\nabla_{\bdelta}D_i(\bdelta)
=-\sum_a p_i(a)\nabla\ell_{i,a}(\bm\theta_k+\bdelta).
\end{equation}
At the anchor,
\begin{equation}
\label{eq:app_score_expectation}
\sum_a p_i(a)\score_{i,a}
=\sum_a \pi_{\bm\theta_k}(a\mid x_i)
\nabla\log\pi_{\bm\theta_k}(a\mid x_i)
=\sum_a\nabla\pi_{\bm\theta_k}(a\mid x_i)
=\nabla 1
=\bm0.
\end{equation}
Therefore
\begin{equation}
\label{eq:app_kl_gradient_zero}
\nabla D_i(\bm0)=\bm0.
\end{equation}

Differentiate the normalization identity once more. For each action,
\begin{equation}
\label{eq:app_prob_hessian}
\nabla^2\pi_{\bm\theta}(a\mid x_i)
=
\pi_{\bm\theta}(a\mid x_i)
\left[
\nabla^2\log\pi_{\bm\theta}(a\mid x_i)
+
\nabla\log\pi_{\bm\theta}(a\mid x_i)
\nabla\log\pi_{\bm\theta}(a\mid x_i)^{\top}
\right].
\end{equation}
Summing over $a$ gives zero because $\sum_a\pi_{\bm\theta}(a\mid x_i)=1$. Evaluated at $\bm\theta_k$,
\begin{equation}
\label{eq:app_fisher_identity}
-\sum_a p_i(a)\nabla^2\ell_{i,a}(\bm\theta_k)
=
\sum_a p_i(a)\score_{i,a}\score_{i,a}^{\top}.
\end{equation}
On the other hand, differentiating Eq.~(\ref{eq:app_kl_gradient}) gives
\begin{equation}
\label{eq:app_kl_hessian_single}
\nabla^2D_i(\bm0)
=-\sum_a p_i(a)\nabla^2\ell_{i,a}(\bm\theta_k).
\end{equation}
Combining Eqs.~(\ref{eq:app_fisher_identity}) and~(\ref{eq:app_kl_hessian_single}),
\begin{equation}
\label{eq:app_single_fisher_hessian}
\nabla^2D_i(\bm0)
=
\mathbb E_{a\sim\pi_{\bm\theta_k}(\cdot\mid x_i)}
[\score_{i,a}\score_{i,a}^{\top}].
\end{equation}
Averaging over $i$ proves Eq.~(\ref{eq:output_hessian}). This is the standard Fisher--KL identity \citep{amari1998natural,martens2020new}.

\subsection{Explicit KL Taylor remainder and mixed derivative}
\label{app:kl_remainder}

Apply multivariate Taylor expansion to $\mathcal D_k$ at $\bm0$. Since its value and gradient vanish,
\begin{equation}
\label{eq:app_kl_taylor}
\mathcal D_k(\bdelta)
=
\frac12\bdelta^{\top}\nabla^2\mathcal D_k(\bm0)\bdelta
+R_{\mathrm{KL}}(\bdelta).
\end{equation}
Under the third-derivative bound in Eq.~(\ref{eq:app_regular_bounds}), the integral or Lagrange remainder satisfies
\begin{equation}
\label{eq:app_kl_remainder_bound}
|R_{\mathrm{KL}}(\bdelta)|
\leq
\frac{M_{\mathcal D}}{6}\lVert\bdelta\rVert^3
\end{equation}
for a local constant $M_{\mathcal D}$. Substituting Eq.~(\ref{eq:output_hessian}) into Eq.~(\ref{eq:app_kl_taylor}) proves the local expansion stated in Lemma~\ref{lem:output_hessian}. The cubic order follows from local Hessian Lipschitzness or, equivalently here, a locally bounded third derivative \citep{nesterov2013introductory,nocedal2006numerical}.

For two directions $\bdelta_1$ and $\bdelta_2$, define $Q(s,t):=\mathcal D_k(s\bdelta_1+t\bdelta_2)$. The chain rule gives
\begin{equation}
\label{eq:app_mixed_kl_derivative}
\left.\frac{\partial^2Q}{\partial s\partial t}\right|_{s=t=0}
=\bdelta_1^{\top}\Fbar_k\bdelta_2.
\end{equation}
Equivalently, polarization of the Fisher quadratic form yields
\begin{equation}
\label{eq:app_fisher_polarization}
2\bdelta_1^{\top}\Fbar_k\bdelta_2
=
\lVert\bdelta_1+\bdelta_2\rVert_{\Fbar_k}^2
-\lVert\bdelta_1\rVert_{\Fbar_k}^2
-\lVert\bdelta_2\rVert_{\Fbar_k}^2,
\end{equation}
where $\lVert\bdelta\rVert_{\Fbar_k}^2:=\bdelta^{\top}\Fbar_k\bdelta$.

\section{Checkpoint Observability of Mixed Output Interaction}
\label{app:checkpoint_observability}

This section derives the preceding and current base footprints on the same realized token coordinates and proves that their product recovers the mixed output interaction under the empirical output Fisher up to cubic local error.

\subsection{Taylor expansion of the preceding checkpoint footprint}
\label{app:first_footprint}

Recall $\bm\theta_{k-1}=\bm\theta_k-\bdelta_{k-1}$. For each realized coordinate,
\begin{equation}
\label{eq:app_u_definition}
u_i=\ell_i(\bm\theta_k)-\ell_i(\bm\theta_k-\bdelta_{k-1}).
\end{equation}
Taylor expansion of $\ell_i(\bm\theta_k-\bdelta_{k-1})$ around $\bm\theta_k$ gives
\begin{equation}
\label{eq:app_backward_taylor}
\ell_i(\bm\theta_k-\bdelta_{k-1})
=
\ell_i(\bm\theta_k)
-\score_i^{\top}\bdelta_{k-1}
+\frac12\bdelta_{k-1}^{\top}
\nabla^2\ell_i(\bm\theta_k-t_i\bdelta_{k-1})
\bdelta_{k-1}
\end{equation}
for some $t_i\in(0,1)$. Hence
\begin{equation}
\label{eq:app_u_expansion}
u_i=\score_i^{\top}\bdelta_{k-1}+e_i^{u},
\end{equation}
where
\begin{equation}
\label{eq:app_u_error}
e_i^{u}
:=-\frac12\bdelta_{k-1}^{\top}
\nabla^2\ell_i(\bm\theta_k-t_i\bdelta_{k-1})
\bdelta_{k-1},
\qquad
|e_i^{u}|\leq\frac{L}{2}\lVert\bdelta_{k-1}\rVert^2.
\end{equation}
In vector form,
\begin{equation}
\label{eq:app_u_vector}
\bm u=\Phimat_k^{\top}\bdelta_{k-1}+\bm e^{u}.
\end{equation}

\subsection{Taylor expansion of the current base footprint}
\label{app:second_footprint}

The current base update ends at $\bm\theta_{k+1}^{0}=\bm\theta_k+\bdelta_k^{0}$. Therefore
\begin{equation}
\label{eq:app_d_definition}
d_i^{0}=\ell_i(\bm\theta_k+\bdelta_k^{0})-\ell_i(\bm\theta_k).
\end{equation}
Taylor expansion around $\bm\theta_k$ gives
\begin{equation}
\label{eq:app_d_expansion}
d_i^{0}=\score_i^{\top}\bdelta_k^{0}+e_i^{d},
\end{equation}
with
\begin{equation}
\label{eq:app_d_error}
e_i^{d}
:=\frac12(\bdelta_k^{0})^{\top}
\nabla^2\ell_i(\bm\theta_k+t_i'\bdelta_k^{0})
\bdelta_k^{0},
\qquad
|e_i^{d}|\leq\frac{L}{2}\lVert\bdelta_k^{0}\rVert^2.
\end{equation}
Thus
\begin{equation}
\label{eq:app_d_vector}
\bm d^{0}=\Phimat_k^{\top}\bdelta_k^{0}+\bm e^{d}.
\end{equation}

\subsection{Proof of checkpoint token--Fisher duality}
\label{app:token_duality}

Multiplying Eqs.~(\ref{eq:app_u_expansion}) and~(\ref{eq:app_d_expansion}) coordinatewise gives
\begin{equation}
\label{eq:app_product_expand}
\begin{aligned}
u_i d_i^{0}
={}&(\score_i^{\top}\bdelta_{k-1})(\score_i^{\top}\bdelta_k^{0})\\
&+(\score_i^{\top}\bdelta_{k-1})e_i^{d}
+(\score_i^{\top}\bdelta_k^{0})e_i^{u}
+e_i^{u}e_i^{d}.
\end{aligned}
\end{equation}
Averaging the leading term,
\begin{equation}
\label{eq:app_leading_product}
\frac1N\sum_{i=1}^{N}
(\score_i^{\top}\bdelta_{k-1})(\score_i^{\top}\bdelta_k^{0})
=
\bdelta_{k-1}^{\top}
\left(\frac1N\sum_{i=1}^{N}\score_i\score_i^{\top}\right)
\bdelta_k^{0}
=
\bdelta_{k-1}^{\top}\Fmat_k\bdelta_k^{0}.
\end{equation}
Using $|\score_i^{\top}\bdelta|\leq G\lVert\bdelta\rVert$ and Eqs.~(\ref{eq:app_u_error})--(\ref{eq:app_d_error}), the total error satisfies
\begin{equation}
\label{eq:app_duality_error}
\begin{aligned}
\left|
\frac1N\bm u^{\top}\bm d^{0}
-\bdelta_{k-1}^{\top}\Fmat_k\bdelta_k^{0}
\right|
\leq{}&
\frac{GL}{2}\lVert\bdelta_{k-1}\rVert\lVert\bdelta_k^{0}\rVert^2\\
&+\frac{GL}{2}\lVert\bdelta_k^{0}\rVert\lVert\bdelta_{k-1}\rVert^2\\
&+\frac{L^2}{4}\lVert\bdelta_{k-1}\rVert^2\lVert\bdelta_k^{0}\rVert^2.
\end{aligned}
\end{equation}
Under Eq.~(\ref{eq:app_small_steps}), the first two terms are $O(\eta^3)$ and the last is $O(\eta^4)$, proving Theorem~\ref{thm:cross_step}. Equation~(\ref{eq:app_duality_error}) also states the finite-step approximation explicitly in addition to the asymptotic form.

\subsection{Population and empirical Fisher}
\label{app:empirical_fisher}

The checkpoint identity is deterministic in the realized empirical output metric $\Fmat_k$, while the Hessian of the full action-distribution KL is $\Fbar_k$. This distinction is important because empirical Fisher matrices should not be identified with arbitrary objective Hessians \citep{kunstner2019limitations}.

If each realized $a_i$ is sampled from the anchor policy conditionally on $x_i$, then
\begin{equation}
\label{eq:app_empirical_unbiased}
\mathbb E[\Fmat_k\mid x_1,\ldots,x_N]
=\Fbar_k.
\end{equation}
For fixed directions $\bdelta_1,\bdelta_2$, define
\begin{equation}
\label{eq:app_directional_variable}
Z_i:=(\score_i^{\top}\bdelta_1)(\score_i^{\top}\bdelta_2).
\end{equation}
Let
\begin{equation}
\label{eq:app_coordinate_fisher}
\Fbar_{k,i}
:=\mathbb E_{a\sim\pi_{\bm\theta_k}(\cdot\mid x_i)}
[\score_{i,a}\score_{i,a}^{\top}].
\end{equation}
Then
\begin{equation}
\label{eq:app_directional_fisher}
\bdelta_1^{\top}\Fmat_k\bdelta_2
=\frac1N\sum_{i=1}^{N}Z_i,
\qquad
\mathbb E[Z_i\mid x_i]
=\bdelta_1^{\top}\Fbar_{k,i}\bdelta_2.
\end{equation}
If $\lVert\score_i\rVert\leq G$, then
\begin{equation}
\label{eq:app_directional_bound}
|Z_i|\leq G^2\lVert\bdelta_1\rVert\lVert\bdelta_2\rVert.
\end{equation}
Under conditional independence, Hoeffding's inequality \citep{hoeffding1963probability} gives, with probability at least $1-\delta$,
\begin{equation}
\label{eq:app_hoeffding}
\left|
\bdelta_1^{\top}(\Fmat_k-\Fbar_k)\bdelta_2
\right|
\leq
G^2\lVert\bdelta_1\rVert\lVert\bdelta_2\rVert
\sqrt{\frac{2\log(2/\delta)}{N}}.
\end{equation}
Combining Eqs.~(\ref{eq:app_duality_error}) and~(\ref{eq:app_hoeffding}) yields
\begin{equation}
\label{eq:app_population_total_error}
\frac1N\bm u^{\top}\bm d^{0}
=
\bdelta_{k-1}^{\top}\Fbar_k\bdelta_k^{0}
+O(\eta^3)
+O_p\!\left(\frac{\eta^2}{\sqrt N}\right).
\end{equation}
For autoregressive trajectories, the deterministic empirical identity remains unchanged; only the optional population concentration argument is applied at the trajectory level or through an appropriate martingale bound.

\subsection{Path cancellation and token-level backtracking}
\label{app:path_cancellation}

For arbitrary vectors $\bm u,\bm d\in\mathbb R^N$,
\begin{equation}
\label{eq:app_norm_expand}
\lVert\bm u+\bm d\rVert^2
=\lVert\bm u\rVert^2+\lVert\bm d\rVert^2+2\bm u^{\top}\bm d.
\end{equation}
Rearranging proves Eq.~(\ref{eq:gamma_path}). Its left-hand side is the sum of the two stepwise squared path lengths minus the squared net displacement, divided by two; it is positive when the two output motions exhibit net cancellation.

For the coordinatewise decomposition, define
\begin{equation}
\label{eq:app_align_rev}
\mathrm{Align}(\bm u,\bm d):=\frac1N\sum_i[u_i d_i]_+,
\qquad
\mathrm{Rev}(\bm u,\bm d):=\frac1N\sum_i[-u_i d_i]_+.
\end{equation}
Since every scalar $x$ satisfies $x=[x]_+-[-x]_+$,
\begin{equation}
\label{eq:app_align_rev_decomp}
\frac1N\bm u^{\top}\bm d
=\mathrm{Align}(\bm u,\bm d)-\mathrm{Rev}(\bm u,\bm d),
\qquad
\Gamma(\bm u,\bm d)=\mathrm{Rev}(\bm u,\bm d)-\mathrm{Align}(\bm u,\bm d).
\end{equation}
Therefore
\begin{equation}
\label{eq:app_global_rev_bound}
[\Gamma(\bm u,\bm d)]_+
=[\mathrm{Rev}-\mathrm{Align}]_+
\leq\mathrm{Rev}.
\end{equation}
Token-level backtracking avoids cancellation between aligned and opposing coordinates. It decomposes into the two sign components
\begin{equation}
\label{eq:app_reversal_branches}
\mathrm{Rev}(\bm u,\bm d)
=
\underbrace{\frac1N\sum_{i:u_i<0}(-u_i)[d_i]_+}_{\mathcal B^-(\bm u,\bm d)}
+
\underbrace{\frac1N\sum_{i:u_i>0}u_i[-d_i]_+}_{\mathcal B^+(\bm u,\bm d)}.
\end{equation}
The first component is the negative-to-positive rebound controlled by \method.

\section{From Residual Coefficients to Backtracking Control}
\label{app:residual_control}

This section proves the control chain used in the main text: the constructed residual remains nonpositive after coefficient clipping, acts as a PPO likelihood-ratio penalty, induces a correction footprint through the token kernel, and contracts the targeted backtracking component.

\subsection{Effective coefficient correction after clipping}
\label{app:advantage_clipping}

The method constructs $r_i^{\methodmath}\leq0$ on $\mathcal C_k$ and $r_i^{\methodmath}=0$ outside it; the complete rank and scale construction is given in Appendix~\ref{app:code_construction}. Set
\begin{equation}
\label{eq:app_clipped_coefficients}
\begin{aligned}
A_i^0
&:=\mathrm{clip}(A_i^{\mathrm{base}},-A_{\max},A_{\max}),\\
\widetilde A_i
&:=\mathrm{clip}(A_i^{\mathrm{base}}+r_i^{\methodmath},-A_{\max},A_{\max}),
\qquad
\xi_i:=\widetilde A_i-A_i^0.
\end{aligned}
\end{equation}
The scalar clipping map is monotone nondecreasing. Hence
\begin{equation}
\label{eq:app_effective_sign}
\xi_i\leq0.
\end{equation}
It is also $1$-Lipschitz, so
\begin{equation}
\label{eq:app_effective_magnitude}
|\xi_i|\leq|r_i^{\methodmath}|.
\end{equation}
Final coefficient clipping therefore never reverses the penalty direction and can only attenuate its magnitude. If $r_i^{\methodmath}=0$, then $\xi_i=0$.

\subsection{Proof of the history-conditioned PPO penalty}
\label{app:ppo_penalty}

Define the token log-ratio and likelihood ratio
\begin{equation}
\label{eq:app_ratio}
z_i(\bm\theta)
:=\log\frac{\pi_{\bm\theta}(a_i\mid x_i)}
{\pi_{\bm\theta_k}(a_i\mid x_i)},
\qquad
\omega_i(\bm\theta):=e^{z_i(\bm\theta)}.
\end{equation}
At the anchor, $z_i=0$ and $\omega_i=1$, which lies in the interior of the PPO clipping interval. In a local neighborhood, the token losses therefore reduce to the unclipped likelihood-ratio form \citep{schulman2017proximal}:
\begin{equation}
\label{eq:app_local_token_losses}
\mathcal J_i^0(z_i)=-e^{z_i}A_i^0,
\qquad
\mathcal J_i^{\methodmath}(z_i)=-e^{z_i}(A_i^0+\xi_i).
\end{equation}
Subtracting,
\begin{equation}
\label{eq:app_penalty_difference}
\mathcal J_i^{\methodmath}(z_i)-\mathcal J_i^0(z_i)
=-e^{z_i}\xi_i
=e^{z_i}|\xi_i|
\geq0.
\end{equation}
Differentiating with respect to $z_i$ and evaluating at the anchor gives
\begin{equation}
\label{eq:app_penalty_derivative}
\left.
\frac{\partial(\mathcal J_i^{\methodmath}-\mathcal J_i^0)}{\partial z_i}
\right|_{z_i=0}
=-\xi_i\geq0,
\end{equation}
which proves Lemma~\ref{lem:history_penalty}. Gradient descent therefore adds a penalty against increasing $z_i$ on the selected support. The result is independent of the sign of $A_i^0$ because it concerns the incremental term induced by $\xi_i$.

\subsection{Base update and correction footprint}
\label{app:mixed_update}

Partition the response-token score matrix into focus and non-focus blocks, $\Phimat_{\mathrm f}$ and $\Phimat_{\mathrm{nf}}$. In the local unclipped policy-gradient model \citep{sutton1999policy,schulman2017proximal}, the base mixed update is
\begin{equation}
\label{eq:app_mixed_base_update}
\bdelta_k^{0}
=\frac{\eta}{|\mathcal R|}
\left(
\Phimat_{\mathrm f}\bm A_{\mathrm f}^{0}
+\Phimat_{\mathrm{nf}}\bm A_{\mathrm{nf}}^{0}
\right).
\end{equation}
Let $\Phimat_{\mathcal E}$ contain the scores of eligible focus coordinates. Their first-order base footprint is
\begin{equation}
\label{eq:app_mixed_base_footprint}
\begin{aligned}
\bm d^{0}
&=\Phimat_{\mathcal E}^{\top}\bdelta_k^{0}+O(\eta^2)\\
&=\frac{\eta}{|\mathcal R|}
\left(
\Phimat_{\mathcal E}^{\top}\Phimat_{\mathrm f}\bm A_{\mathrm f}^{0}
+\Phimat_{\mathcal E}^{\top}\Phimat_{\mathrm{nf}}\bm A_{\mathrm{nf}}^{0}
\right)
+O(\eta^2).
\end{aligned}
\end{equation}
Thus $\bm d^0$ already includes the complete focus- and non-focus-domain contribution to the base update; the control argument does not remove or isolate non-focus contributions.

The effective \method\ correction is supported on eligible focus tokens:
\begin{equation}
\label{eq:app_residual_update}
\bdelta_k^{r}
=\frac{\eta}{|\mathcal R|}\Phimat_{\mathcal E}\bxi.
\end{equation}
Its correction footprint is
\begin{equation}
\label{eq:app_residual_footprint}
\bh
:=\Phimat_{\mathcal E}^{\top}\bdelta_k^{r}
=\frac{\eta}{|\mathcal R|}\Kmat_{\mathcal E}\bxi,
\qquad
\Kmat_{\mathcal E}:=\Phimat_{\mathcal E}^{\top}\Phimat_{\mathcal E}.
\end{equation}
This is the standard first-order function-space motion induced by a Jacobian Gram matrix \citep{jacot2018neural}. Freezing the optimizer preconditioner at iteration $k$ as a positive-semidefinite matrix $P_k$ gives the local preconditioned response
\begin{equation}
\label{eq:app_preconditioned_kernel}
\bh
=\frac{\eta}{|\mathcal R|}
\Phimat_{\mathcal E}^{\top}P_k\Phimat_{\mathcal E}\bxi.
\end{equation}
The preconditioned token kernel remains positive semidefinite. For AdamW, $P_k$ approximates the local change in the optimizer update with its moment state frozen at iteration $k$; together with the unclipped PPO model at the anchor, this captures the first-order effect of the correction rather than the full clipped PPO--AdamW trajectory.

\subsection{Residual-to-output sign transfer}
\label{app:self_response}

From Eq.~(\ref{eq:app_residual_footprint}), for $i\in\mathcal C_k$,
\begin{equation}
\label{eq:app_h_coordinate}
h_i
=\frac{\eta}{|\mathcal R|}
\left[
(\Kmat_{\mathcal E})_{ii}\xi_i
+\sum_{j\in\mathcal C_k,\,j\neq i}
(\Kmat_{\mathcal E})_{ij}\xi_j
\right].
\end{equation}
A sufficient local self-response condition is
\begin{equation}
\label{eq:app_self_response_condition}
(\Kmat_{\mathcal E})_{ii}|\xi_i|
\geq
\left|
\sum_{j\in\mathcal C_k,\,j\neq i}
(\Kmat_{\mathcal E})_{ij}\xi_j
\right|.
\end{equation}
Since $(\Kmat_{\mathcal E})_{ii}=\lVert\score_i\rVert^2\geq0$ and $\xi_i\leq0$, the diagonal term is nonpositive. Under Eq.~(\ref{eq:app_self_response_condition}),
\begin{equation}
\label{eq:app_sign_transfer_bound}
\begin{aligned}
(\Kmat_{\mathcal E})_{ii}\xi_i
+\sum_{j\neq i}(\Kmat_{\mathcal E})_{ij}\xi_j
&\leq
-(\Kmat_{\mathcal E})_{ii}|\xi_i|
+
\left|\sum_{j\neq i}(\Kmat_{\mathcal E})_{ij}\xi_j\right|\\
&\leq0,
\end{aligned}
\end{equation}
which proves Lemma~\ref{lem:sign_transfer}. The condition holds automatically for a diagonal token kernel and under weighted diagonal dominance on the selected coordinates.

Coordinatewise nonpositivity is sufficient but not necessary. Using
\begin{equation}
\label{eq:app_positive_part_integral}
[x+y]_+-[x]_+
=\int_0^1\mathbf 1\{x+ty>0\}\,y\,dt
\end{equation}
for almost every $(x,y)$, the exact first-order backtracking change is
\begin{equation}
\label{eq:app_aggregate_change}
\begin{aligned}
&\mathcal B^{-}(\bm u,\bm d^0+\bh)
-\mathcal B^{-}(\bm u,\bm d^0)\\
&\quad=\frac1N\int_0^1
\sum_{i\in\mathcal C_k}
(-u_i)\mathbf 1\{d_i^0+t h_i>0\}h_i\,dt.
\end{aligned}
\end{equation}
Hence the weaker aggregate condition
\begin{equation}
\label{eq:app_aggregate_condition}
\sum_{i\in\mathcal C_k}
(-u_i)\mathbf 1\{d_i^0+t h_i>0\}h_i\leq0,
\qquad t\in[0,1],
\end{equation}
is sufficient for contraction even if a small number of individual $h_i$ are positive because of cross-token coupling.

\subsection{Proof of first-order backtracking contraction}
\label{app:backtracking_proof}

For $i\in\mathcal C_k$, $u_i<0$. Under $h_i\leq0$, monotonicity of the positive-part function gives
\begin{equation}
\label{eq:app_positive_part_contract}
[d_i^0+h_i]_+\leq[d_i^0]_+.
\end{equation}
Multiplying by $-u_i>0$, summing, and dividing by $N$ yields
\begin{equation}
\label{eq:app_backtracking_contract}
\mathcal B^{-}(\bm u,\bm d^0+\bh)
\leq\mathcal B^{-}(\bm u,\bm d^0).
\end{equation}
The nonnegative difference is exactly the control gain in Eq.~(\ref{eq:control_gain}). For the selected-coordinate signed interaction,
\begin{equation}
\label{eq:app_alignment_gain}
\begin{aligned}
&\frac1N\sum_{i\in\mathcal C_k}u_i(d_i^0+h_i)
-\frac1N\sum_{i\in\mathcal C_k}u_i d_i^0\\
&\qquad=\frac1N\sum_{i\in\mathcal C_k}u_i h_i\geq0,
\end{aligned}
\end{equation}
because both factors are nonpositive. Strict inequality holds whenever $u_i<0$ and $h_i<0$ on at least one selected coordinate.

\subsection{Finite-step control gain}
\label{app:finite_control}

The local proof uses first-order footprints. The corrected footprint is
\begin{equation}
\label{eq:app_finite_corrected_footprint}
\bm d^{\methodmath}
:=\bm\ell_k(\bm\theta_k+\bdelta_k^0+\bdelta_k^r)
-\bm\ell_k(\bm\theta_k).
\end{equation}
Taylor expansion around $\bm\theta_k$ gives
\begin{equation}
\label{eq:app_finite_footprint_expand}
\bm d^{\methodmath}
=\bm d^0+\bh+\bm e^r,
\qquad
\lVert\bm e^r\rVert_\infty=O(\eta^2).
\end{equation}
Since $x\mapsto[x]_+$ is $1$-Lipschitz,
\begin{equation}
\label{eq:app_finite_lipschitz}
\begin{aligned}
&\left|
\mathcal B^-(\bm u,\bm d^{\methodmath})
-\mathcal B^-(\bm u,\bm d^0+\bh)
\right|\\
&\quad\leq
\frac1N\sum_{i\in\mathcal C_k}(-u_i)|e_i^r|
\leq
\frac{\lVert\bm u\rVert_1}{N}\lVert\bm e^r\rVert_\infty
=O(\eta^3),
\end{aligned}
\end{equation}
because $\lVert\bm u\rVert_1/N=O(\eta)$. Therefore
\begin{equation}
\label{eq:app_finite_contract_error}
\begin{aligned}
\mathcal B^-(\bm u,\bm d^{\methodmath})
&\leq
\mathcal B^-(\bm u,\bm d^0+\bh)+O(\eta^3)\\
&=
\mathcal B^-(\bm u,\bm d^0)
-\Delta_k^{\mathrm{ctrl}}
+O(\eta^3),
\end{aligned}
\end{equation}
which proves Eq.~(\ref{eq:finite_control_gain_main}). This form retains the controller's first-order reduction explicitly and separates it from the cubic deviation incurred by observing a finite checkpoint step.

\section{Design Principles and Code-Faithful Construction}
\label{app:design_principles}

The preceding sections establish how a nonpositive effective coefficient becomes a backtracking-reducing output correction. This section derives the three design choices used to construct that coefficient: penalty-only support, rank-based allocation, and batchwise scale normalization.

\subsection{Why \method\ uses the penalty-only support}
\label{app:penalty_asymmetry}

The full tokenwise backtracking statistic is symmetric in the sign patterns $u_i<0,d_i>0$ and $u_i>0,d_i<0$. A symmetric controller would apply a nonpositive correction after negative drift and a nonnegative correction after positive drift. \method\ restricts the correction to the negative-drift candidate set because a nonpositive correction has a monotone backtracking-reduction property there and does not introduce a positive reward based only on checkpoint history.

For an arbitrary nonpositive output-space correction $v_i\leq0$, define
\begin{equation}
\label{eq:app_scalar_reversal}
b_i(v_i):=[-u_i(d_i+v_i)]_+.
\end{equation}
If $u_i<0$,
\begin{equation}
\label{eq:app_negative_support_monotone}
b_i(v_i)
=(-u_i)[d_i+v_i]_+
\leq(-u_i)[d_i]_+
=b_i(0).
\end{equation}
If $u_i>0$,
\begin{equation}
\label{eq:app_positive_support_monotone}
b_i(v_i)
=u_i[-d_i-v_i]_+
\geq u_i[-d_i]_+
=b_i(0).
\end{equation}
Thus a nonpositive correction is backtracking-reducing on $\{u_i<0\}$ but can increase backtracking on $\{u_i>0\}$.

For any $\bm v\preceq0$, define its negative-drift projection
\begin{equation}
\label{eq:app_projected_correction}
v_i^{-}:=
\begin{cases}
v_i,&u_i<0,\\
0,&u_i\geq0.
\end{cases}
\end{equation}
Applying Eqs.~(\ref{eq:app_negative_support_monotone})--(\ref{eq:app_positive_support_monotone}) coordinatewise gives
\begin{equation}
\label{eq:app_safe_support_projection}
\mathrm{Rev}(\bm u,\bm d+\bm v^{-})
\leq
\mathrm{Rev}(\bm u,\bm d+\bm v).
\end{equation}
This identifies $\{u_i<0\}$ as the support on which a nonpositive correction cannot increase backtracking; Lemma~\ref{lem:sign_transfer} connects this support to the implemented coefficient correction.

Suppressing the complementary pattern $u_i>0,d_i<0$ would require a positive coefficient correction $\xi_i^{+}>0$. At the PPO anchor,
\begin{equation}
\label{eq:app_positive_reward_signal}
\mathcal J_i^{+}-\mathcal J_i^0=-e^{z_i}\xi_i^{+},
\qquad
\left.
\frac{\partial(\mathcal J_i^{+}-\mathcal J_i^0)}{\partial z_i}
\right|_{z_i=0}
=-\xi_i^{+}<0,
\end{equation}
so gradient descent treats the historical signal as an additional reward-like incentive. This changes the role of history in four ways.

\paragraph{Historical-drift self-reinforcement.}
A preceding positive movement would itself trigger an additional positive coefficient, making the same direction more likely to continue even when the new rollout provides no confirming reward evidence. A transient checkpoint displacement can thereby become a persistent optimization preference.

\paragraph{Positive feedback on sampling noise.}
The measured drift contains finite-rollout, optimizer, and checkpoint variability. A positive residual maps a noisy positive realization into another positive update, creating a feedback loop that amplifies rather than damps the historical fluctuation.

\paragraph{Persistent elevation of incorrect tokens.}
A token probability can rise because of incidental generalization, cross-token coupling, or an update that benefits another domain. Historical sign alone does not certify correctness; rewarding every previous increase can repeatedly elevate a token whose current reward evidence is weak or adverse.

\paragraph{Conflict with the current GRPO advantage.}
When $A_i^{\mathrm{base}}<0$, a positive history residual cancels part of the current reward-grounded negative coefficient and can even reverse its sign. The update would then prefer historical motion over the latest within-group reward comparison.

This penalty-only design avoids introducing a new positive incentive solely from checkpoint history. It can attenuate a likelihood increase, but it cannot create a positive incentive absent from the current GRPO coefficient. This asymmetry is therefore a control design rather than an omission: \method\ uses the candidate set with a monotone backtracking-reduction guarantee while preserving the current reward as the sole source of positive reinforcement.

\subsection{Rank-optimal local allocation}
\label{app:rank_allocation}

Write a nonpositive output correction as $v_i=-e_i$ with $e_i\geq0$. For $u_i<0$, the exact reduction in scalar backtracking is
\begin{equation}
\label{eq:app_scalar_reduction}
\begin{aligned}
[-u_i d_i]_+-[-u_i(d_i-e_i)]_+
&=(-u_i)\left([d_i]_+-[d_i-e_i]_+\right)\\
&=(-u_i)\min\{e_i,[d_i]_+\}.
\end{aligned}
\end{equation}
When $d_i>0$ and $0\leq e_i\leq d_i$, the reduction is linear:
\begin{equation}
\label{eq:app_active_linear_reduction}
[-u_i d_i]_+-[-u_i(d_i-e_i)]_+
=(-u_i)e_i,
\end{equation}
so the marginal reduction per unit penalty is $-u_i$.

\begin{proposition}[Rank-optimal local allocation]
\label{prop:rank_allocation}
Consider active rebound coordinates with $u_i<0$ and $d_i>0$. Let $\{e_i\}$ be a fixed multiset of unsaturated penalty levels satisfying $0\leq e_i\leq\min_j d_j$, so every level is feasible on every active coordinate. The total local reduction
\begin{equation}
\label{eq:app_total_rank_reduction}
\Delta\mathcal B^-
=\frac1N\sum_i(-u_i)e_i
\end{equation}
is maximized by assigning larger $e_i$ to more negative $u_i$.
\end{proposition}

\paragraph{Proof.}
Let $-u_i\geq-u_j$ but $e_i<e_j$. Swapping the two penalty levels changes the objective by
\begin{equation}
\label{eq:app_pairwise_swap}
\frac1N
\left[
(-u_i)e_j+(-u_j)e_i
-(-u_i)e_i-(-u_j)e_j
\right]
=
\frac1N\bigl((-u_i)-(-u_j)\bigr)(e_j-e_i)
\geq0.
\end{equation}
Repeatedly eliminating such inversions produces a co-monotone assignment without decreasing the reduction. Hence stronger penalties should be assigned to more negative preceding drift. \hfill$\square$

Proposition~\ref{prop:rank_allocation} supplies the allocation principle behind Eq.~(\ref{eq:m_quant}). The quantile map implements the optimal ordering without using raw drift magnitude, is robust to isolated extremes, and separates relative allocation from the common scale chosen by $\tau$. Under the diagonal self-response approximation $h_i\simeq\alpha_i\xi_i$ with positive, locally comparable $\alpha_i$, the same ordering transfers from coefficient penalties to the correction footprint.

\subsection{Code-faithful residual construction}
\label{app:code_construction}

The base coefficient is
\begin{equation}
\label{eq:app_base_advantage}
A_i^{\mathrm{base}}=c_iA_i^{\mathrm g},
\qquad
c_i=\frac{w_{\mathrm{src}(i)}}{Z_k},
\end{equation}
where $Z_k$ is given by Eq.~(\ref{eq:m_taskw}). The history residual is not multiplied by $c_i$ again. The eligible set $\mathcal E_k$ contains response-token coordinates from the current focus domain whose trajectory-level GRPO advantage is nonzero, and
\begin{equation}
\label{eq:app_candidate_set}
\mathcal C_k=\{i\in\mathcal E_k:u_i<0\}.
\end{equation}

For $n_k:=|\mathcal C_k|>0$, sort candidate drifts in ascending order and assign
\begin{equation}
\label{eq:app_percentile}
p_i=\frac{\mathrm{rank}_0(u_i)+1/2}{n_k}.
\end{equation}
The clipped affine profile is
\begin{equation}
\label{eq:app_raw_residual}
\bar r_i
=r_{\min}
+\frac{\mathrm{clip}(p_i,p_{\mathrm{lo}},p_{\mathrm{hi}})-p_{\mathrm{lo}}}
{p_{\mathrm{hi}}-p_{\mathrm{lo}}}
(r_{\max}-r_{\min}),
\end{equation}
with $\bar r_i=0$ off $\mathcal C_k$. Since ascending rank is monotone,
\begin{equation}
\label{eq:app_raw_monotone}
u_i\leq u_j<0
\quad\Longrightarrow\quad
\bar r_i\leq\bar r_j\leq0.
\end{equation}
For $(p_{\mathrm{lo}},p_{\mathrm{hi}},r_{\min},r_{\max})=(0.1,1,-1,0)$,
\begin{equation}
\label{eq:app_standard_profile}
\bar r_i=-1+\frac{\mathrm{clip}(p_i,0.1,1)-0.1}{0.9}.
\end{equation}

The implementation retains an explicit global gate $g_k\geq0$ and focus mask $\lambda_i\in[0,1]$:
\begin{equation}
\label{eq:app_effective_raw}
q_i=g_k\lambda_i\bar r_i.
\end{equation}
In the standard configuration, $g_k=1$ and $\lambda_i=1$ on the eligible focus coordinates, so Eq.~(\ref{eq:app_effective_raw}) reduces to the conceptual profile in Eq.~(\ref{eq:m_effective_raw}). The explicit terms are retained in code to support scheduling and masking without changing the core method.

\paragraph{Policy objective.}
After forming $\widetilde A_i$ in Eq.~(\ref{eq:m_adv}), define
\begin{equation}
\label{eq:m_ratio}
\omega_i(\bm\theta)
:=\frac{\pi_{\bm\theta}(a_i\mid x_i)}
{\pi_{\bm\theta_{\mathrm{old}}}(a_i\mid x_i)}.
\end{equation}
The controller uses the unchanged clipped surrogate \citep{schulman2017proximal}:
\begin{equation}
\label{eq:m_obj}
\mathcal L_{\methodmath}(\bm\theta)
=-\mathbb E\!\left[
\frac1{|\mathcal R|}\sum_{i\in\mathcal R}
\min\!\left(
\omega_i(\bm\theta)\widetilde A_i,
\mathrm{clip}(\omega_i(\bm\theta),1-\epsilon_c,1+\epsilon_c)\widetilde A_i
\right)
\right].
\end{equation}
No auxiliary objective is introduced; \method\ changes only the coefficient supplied to the existing policy update.

\subsection{Adaptive scale identity and degenerate cases}
\label{app:quantile_scale}

Let
\begin{equation}
\label{eq:app_kappa}
\kappa_k
=\tau\frac{\sigma_A}{\sigma_q},
\qquad
\sigma_A:=\mathrm{std}_{\mathcal E_k}(A^{\mathrm{base}}),
\qquad
\sigma_q:=\mathrm{std}_{\mathcal E_k}(q).
\end{equation}
When $\sigma_A$ or $\sigma_q$ is nonfinite, or $\sigma_q$ falls below the implementation threshold, the method sets $\kappa_k=0$. Otherwise,
\begin{equation}
\label{eq:app_final_residual}
r_i^{\methodmath}=\kappa_kq_i
=\kappa_kg_k\lambda_i\bar r_i\leq0,
\end{equation}
and homogeneity of standard deviation gives
\begin{equation}
\label{eq:app_scale_proof}
\mathrm{std}_{\mathcal E_k}(r^{\methodmath})
=\kappa_k\mathrm{std}_{\mathcal E_k}(q)
=\tau\mathrm{std}_{\mathcal E_k}(A^{\mathrm{base}}).
\end{equation}
Thus
\begin{equation}
\label{eq:app_tau_budget}
\frac{\mathrm{std}_{\mathcal E_k}(r^{\methodmath})}
{\mathrm{std}_{\mathcal E_k}(A^{\mathrm{base}})}
=\tau.
\end{equation}
The identity is a batchwise scale normalization: $\tau$ fixes residual dispersion relative to base-advantage dispersion. Because noncandidate zeros remain in the eligible-set statistic, a sparse candidate set concentrates that dispersion on fewer coordinates, whereas a dense set distributes it more broadly. If $\mathcal C_k=\varnothing$ or the scale statistics are degenerate, $\kappa_k=0$ and \method\ reduces exactly to the base mixed-domain objective.

\subsection{Main-Training Reproducibility Details}
\label{app:main_training_config}

Table~\ref{tab:main_training_config} summarizes the training, generation, and \method-specific settings used in the main experiments. Let $\mathcal D$ denote the four training domains. At trainer step $k$, the focus domain is sampled uniformly from the domains other than the preceding focus,
\begin{equation}
f_k \sim \operatorname{Unif}\!\left(\mathcal D \setminus \{f_{k-1}\}\right).
\end{equation}
The no-repeat constraint prevents one domain from leading two consecutive updates, while random selection avoids imposing a fixed domain order. All four domains remain in the mixed-domain batch; the sampled focus domain determines the asymmetric base weight and the response tokens eligible for history correction. Each trainer step contains 128 prompts and eight sampled responses per prompt, yielding 1,024 rollout trajectories before optimization. Unless otherwise specified, the optimization and rollout settings follow the standard GRPO configuration in verl v0.5.0.

\begin{table}[t]
\centering
\small
\setlength{\tabcolsep}{5pt}
\renewcommand{\arraystretch}{1.08}
\caption{Main-training and generation settings. The prompt batch size is counted before rollout expansion.}
\label{tab:main_training_config}
\begin{tabular}{p{0.38\linewidth}p{0.56\linewidth}}
\toprule
Setting & Configuration \\
\midrule
Model backbones & Qwen3-30B-A3B and Qwen3-8B-Base \\
Training domains & Mathematics, code generation, instruction following, and open-ended Chat \\
Training examples & 7.5K examples per domain \\
Focus-domain schedule & Uniform random sampling from $\mathcal D \setminus \{f_{k-1}\}$; consecutive steps cannot use the same focus domain \\
Prompt batch size & 128 prompts per trainer step \\
Rollouts per prompt & 8 \\
Rollout trajectories per step & 1,024 \\
Training duration & 15 epochs \\
Optimizer & AdamW, $\beta_1=0.9$, $\beta_2=0.999$, weight decay $0.01$ \\
Learning rate & $1\times10^{-6}$ with a constant schedule and no warmup \\
PPO epochs & 1 per rollout batch \\
PPO clip ratio & $0.2$ (symmetric clipping to $[0.8,1.2]$) \\
Sampling temperature & 1.0 \\
Top-$p$ / Top-$k$ & $1.0$ / $-1$ (no nucleus or top-$k$ truncation) \\
Maximum response length & 8,192 tokens \\
Chat reward model & Skywork-Reward-Llama-3.1-8B-v0.2, following the Chat reward configuration used by MGS \\
\midrule
History checkpoint & The frozen preceding checkpoint $\theta_{k-1}$ \\
Eligible history tokens & Focus-domain response tokens with nonzero trajectory-level GRPO advantage \\
Negative-drift candidates & Tokens satisfying $u_i < 0$ \\
Rank-profile parameters & $(p_{\mathrm{lo}},p_{\mathrm{hi}},r_{\min},r_{\max})=(0.1,1,-1,0)$ \\
Residual scale & $\tau=0.03$ \\
Final coefficient range & $[-3,3]$ \\
\bottomrule
\end{tabular}
\end{table}

The preceding checkpoint is used only to rescore the same realized focus-domain token coordinates sampled by the current policy. All checkpoint-derived quantities are treated as fixed coefficients during optimization, and gradients are not propagated through $\theta_{k-1}$. The resulting history residual is added to the task-weighted GRPO coefficient before applying the standard clipped policy objective.

\end{document}